\documentclass[journal]{IEEEtran}

\usepackage{amsmath,amssymb}
\usepackage{fancyhdr}
\usepackage{graphicx}
\usepackage[pagebackref=true,breaklinks=true,letterpaper=true,colorlinks,bookmarks=false]{hyperref}

\renewcommand{\mathbf}{\boldsymbol}

\begin{document}

\title{Deep Multi-View Learning using Neuron-Wise Correlation-Maximizing Regularizers}

\author{Kui~Jia,
        Jiehong~Lin,
        Mingkui~Tan,
        and~Dacheng~Tao
\thanks{K. Jia and J. Lin are with the School of Electronic and Information Engineering, South China University of Technology, Guangzhou, China. Email: kuijia@scut.edu.cn, lin.jiehong@mail.scut.edu.cn.}
\thanks{M. Tan is with the School of Software Engineering, South China University of Technology, Guangzhou, China. Email: mingkuitan@scut.edu.cn.}
\thanks{D. Tao is with the School of Information Technologies, The University of Sydney, New South Wales, Australia. Email: dacheng.tao@sydney.edu.au.}

}

%
%


\maketitle

\begin{abstract}
Many machine learning problems concern with discovering or associating common patterns in data of multiple views or modalities. Multi-view learning is of the methods to achieve such goals. Recent methods propose deep multi-view networks via adaptation of generic Deep Neural Networks (DNNs), which concatenate features of individual views at intermediate network layers (i.e., \emph{fusion layers}). In this work, we study the problem of multi-view learning in such end-to-end networks. We take a regularization approach via multi-view learning criteria, and propose a novel, effective, and efficient neuron-wise correlation-maximizing regularizer. We implement our proposed regularizers collectively as a \emph{correlation-regularized network layer (CorrReg)}. CorrReg can be applied to either fully-connected or convolutional fusion layers, simply by replacing them with their CorrReg counterparts. By partitioning neurons of a hidden layer in generic DNNs into multiple subsets, we also consider a multi-view feature learning perspective of generic DNNs. Such a perspective enables us to study deep multi-view learning in the context of regularized network training, for which we present control experiments of benchmark image classification to show the efficacy of our proposed CorrReg. To investigate how CorrReg is useful for practical multi-view learning problems, we conduct experiments of RGB-D object/scene recognition and multi-view based 3D object recognition, using networks with fusion layers that concatenate intermediate features of individual modalities or views for subsequent classification. Applying CorrReg to fusion layers of these networks consistently improves classification performance. In particular, we achieve the new state of the art on the benchmark RGB-D object and RGB-D scene datasets. We make the implementation of CorrReg publicly available.
\end{abstract}


%
\IEEEpeerreviewmaketitle

\section{Introduction}\label{SecIntro}

\IEEEPARstart{M}{any} machine learning problems concern with discovering or associating common patterns in data of multiple views or modalities. Typical applications include retrieving images from texts or vice versa, combining visual and audio signals for content understanding, and object recognition from visual observations of multiple modalities. Data of different views usually contain complementary information, whose statistical distributions in the high-dimensional measurements of individual views may also be different. Multi-view learning methods aim to exploit information contained in multiple views to better accomplish specified learning tasks. In this work, we take image classification, in particular multi-view or multi-modal object recognition (e.g., recognizing objects from RGB and depth images), as the primary example to study the problem of multi-view learning.

Given feature observations of different views, existing multi-view learning approaches learn latent space representations in either deterministic \cite{CCA,DCCA,MultimodalDL,DeViSE} or probabilistic manners \cite{ProbCCA,MultiModalPLSA,MultiModalLDA,MultiModalDBM}. The learning objective is to make resulting features of different views at each dimension of the latent space more \emph{related} with each other, where relations may be measured by different metrics/criteria \cite{StocOptmPCAPLS,Hardoon2004,DeViSE}. Among various techniques, Canonical Correlation Analysis (CCA) \cite{CCA} and its extensions \cite{KCCA,SparseCCA,DCCA} are the most representative ones. For example, given two-view data, CCA learns pairs of linear projections so that in the projected space, features of both views are maximally correlated at corresponding dimensions.

\begin{figure}[t]
\centering

\includegraphics[scale=0.42]{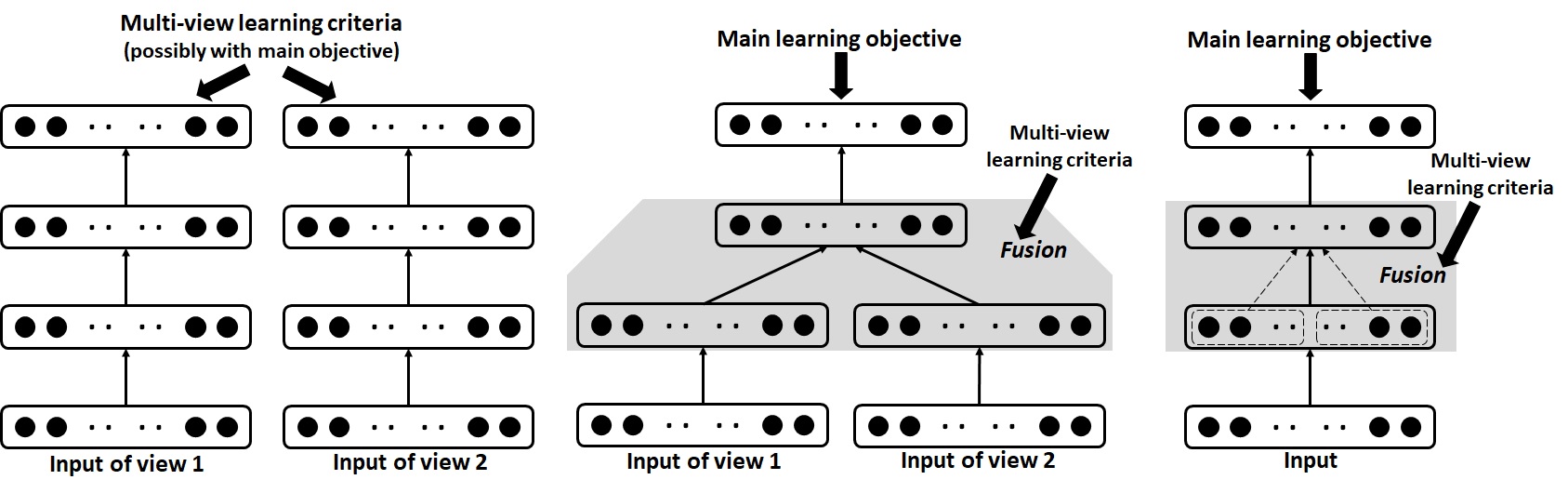} \\

\hfill\hfill\hfill\hfill\hfill\hfill (a) \hfill\hfill\hfill\hfill\hfill\hfill\hfill\hfill\hfill\hfill\hfill\hfill (b)  \hfill\hfill\hfill\hfill\hfill\hfill\hfill\hfill\hfill (c)  \hfill\hfill\hfill\hfill

\caption{Two-view illustrations of deep networks for multi-view learning, where \emph{fusion layers} are inside shaded regions that concatenate features of individual views. (a) Deep multi-view features are learned by applying multi-view learning criteria, possibly together with the main learning objective, on top of multiple single-view deep networks; (b) a deep network takes as inputs data of multiple views/modalities for a specified learning task (e.g., multi-modal image classification), where features of individual views are concatenated at an intermediate layer (i.e., the fusion layer), and multi-view learning criteria can be imposed as a regularization on the fusion layer; (c) in generic DNNs, input neurons of a hidden layer can be partitioned into multiple subsets (represented as black circles grouped in different dashed boxes), and features learned at different subsets could be considered as multi-view features of the same input data.   }  \label{FigMvNetsIllus}
\end{figure}

Following the success of deep learning \cite{AlexNet,RCNNPAMI}, deep multi-view learning methods \cite{DCCA,DeepMultiViewRepLearn} are also proposed recently for learning deep features from multi-view data. These methods apply multi-view learning criteria (e.g., CCA) on top of multiple single-view deep networks (cf. Figure \ref{FigMvNetsIllus}-(a) for an illustration); a two-stage scheme of iterative learning is usually adopted to train the network parameters, where view-specific features are learned until the very top layers, to which either a sequential step of multi-view criteria followed by the objectives of the specified learning tasks or regularized learning objectives that seek a balance between multi-view criteria and the final tasks of interest, are applied. Alternatively, one may design deep architectures that concatenate at intermediate network layers (\emph{fusion layers}) output features of lower, parallel layer streams for individual views \cite{MultimodalDL,MultiModalDBM,ConvRecursive3DObjRecog}, followed by upper network layers for specified learning tasks (e.g., image classification, cf. Figure \ref{FigMvNetsIllus}-(b) for an illustration). Such end-to-end networks have the advantage that the final tasks of interest are achieved directly at the network outputs. However, output features of the lower, parallel streams in such networks capture view-specific patterns, which may not be aligned in a common space for a ready fusion in the subsequent layers.

To enjoy the advantage of end-to-end learning while collaboratively benefiting from different views, multi-view learning criteria could be exploited to improve the relations between resulting features of the lower, parallel streams. Under the framework of regularized function learning, this amounts to training network parameters by penalizing objectives of the main learning tasks with correlation-maximizing regularization at fusion layers (cf. Figure \ref{FigMvNetsIllus}-(b)). In this work, we are interested in CCA criteria since they have long been the main workhorse for multi-view learning \cite{KCCA,Akaho01akernel,Hardoon2004}. Empirical results also show that CCA based approaches outperform alternative ones in the context of deep multi-view representation learning \cite{DeepMultiViewRepLearn}. Directly using CCA as the regularizer makes network training very expensive, which is also incompatible with the mini-batch based stochastic gradient descent (SGD), the commonly used algorithm in deep network training (cf. Section \ref{SecCorrReg} for a discussion). Inspired by batch normalization \cite{BatchNorm}, we propose in this paper a novel {\it neuron-wise correlation-maximizing regularizer}, and implement the proposed regularizers collectively as a \emph{correlation-regularized network layer (CorrReg)}. CorrReg can be applied to either fully-connected (FC) or convolutional (conv) fusion layers, simply by replacing these layers with their CorrReg versions (cf. Figure \ref{FigCorrRegFusionIllus} for an illustration). CorrReg fusion has the same computational complexity as the plain one does, which is significantly lower than that of CCA regularization.


We note that the fusion layer in a multi-view network of Figure \ref{FigMvNetsIllus}-(b) is \emph{computationally} equivalent to any hidden layer in a generic deep neural network (DNN), by partitioning neurons of the hidden layer into multiple subsets (cf. Figure \ref{FigMvNetsIllus}-(c) for an illustration). This equivalence suggests a multi-view feature learning perspective of generic DNNs: when considering hidden neurons of generic DNNs as pattern detectors that characterize different patterns of the input data \cite{ZeilerVisualizing}, features learned at each neuron subset of a hidden layer could be considered as a specific {\it view} of the input data. Ideally, each of such feature views is to learn salient or discriminative patterns of the input data, which should also be generalizable to unseen data. However, there exists an issue of \emph{overfitting}, a phenomenon that specific subsets of layer neurons are trained to be co-adapted to certain patterns in the training data, but cannot generalize well on the held-out test data \cite{Dropout}. As suggested by traditional learning theory \cite{FoundationMLBook}, the risk of overfitting could be severe for generic DNNs, since modern DNNs have large model capacities and are usually over-parameterized in the sense that they can be trained to fit randomly labelled datasets \cite{ZhangICLRBestPaper}. Such a risk for generic DNNs can be addressed implicitly by SGD training \cite{StabilitySGD,DataDependentStabilitySGD,PoggioDLTheoryIIb}, and explicitly by additional regularization \cite{Dropout,BatchNorm}. Our proposed CorReg takes the second regularization approach: improving correlations of neuron subsets reduces co-adaptation to possibly noisy or irrelevant, subset-specific patterns, and thus alleviates the problem of overfitting. Such a connection with generic DNNs enables us to study CorrReg in the context of regularized network training, and to compare with modern regularization techniques (e.g., Dropout \cite{Dropout}) for deep multi-view representation learning.

We finally summarize our contributions as follows.

\begin{itemize}
\item We study in this work the problem of learning deep representations from multi-view data in end-to-end networks. We take a regularization approach via multi-view learning criteria, and propose a novel, effective, and efficient neuron-wise correlation-maximizing regularizer. We implement our proposed regularizers collectively as a correlation-regularized network layer (CorrReg), which will be made publicly available. CorrReg can be applied to either FC or conv based fusion layers, simply by replacing these layers with their CorrReg versions. CorrReg fusion layer has the same computational complexity as a plain one does, which is significantly lower than that of CCA regularization.

\item  We consider a multi-view feature learning perspective of generic DNNs, by partitioning neurons of a hidden layer in a generic DNN into multiple subsets. Such a connection with generic DNNs enables us to study deep multi-view representation learning in the context of regularized network training, and to compare CorrReg with modern regularization techniques (e.g., Dropout \cite{Dropout}). We note that such a comparison is largely ignored in existing deep multi-view learning methods.

\item To investigate the efficacy of CorrReg for regularization of network training, we conduct control experiments on benchmark image classification \cite{Cifar} using generic DNNs \cite{LeCun98,ResNet,WideResNet,DenseNet}. CorrReg consistently improves performance of these networks. To investigate how CorrReg is useful for practical multi-view learning problems, we conduct experiments of RGB-D object/scene recognition and multi-view based 3D object recognition, using networks with fusion layers that concatenate intermediate features of individual modalities or views for subsequent classification. Applying CorrReg to fusion layers of these networks consistently improves classification performance. In particular, we achieve the new state of the art on the benchmark RGB-D object \cite{RGBDObjDataset} and RGB-D scene \cite{SUNdataset} datasets.

\end{itemize}

\section{The Proposed Neuron-Wise Correlation-Maximizing Regularizers}
\label{SecCorrReg}

\begin{figure}[t]
\centering

\includegraphics[scale=0.28]{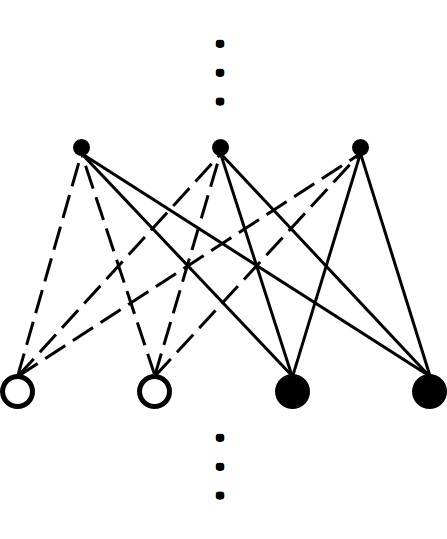} \hfill  \includegraphics[scale=0.28]{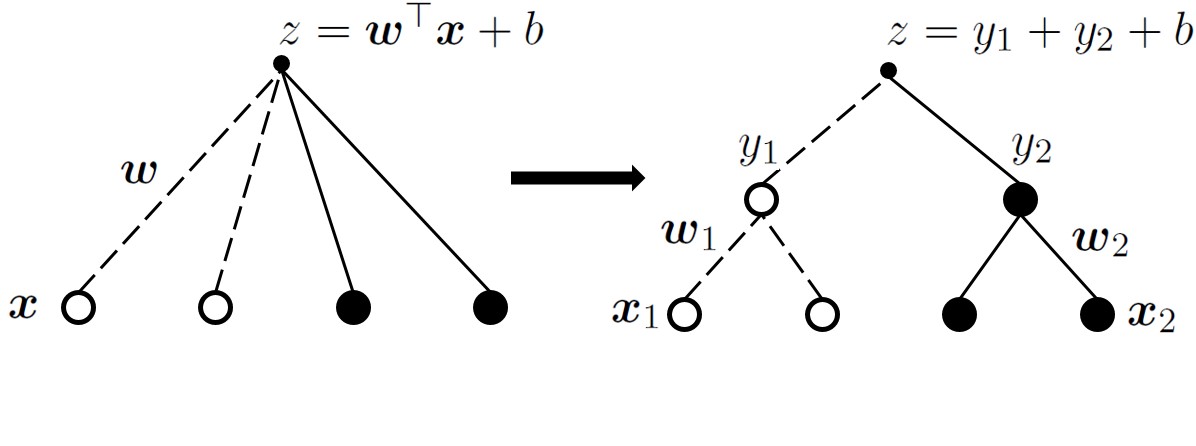} \\

\caption{Two sub-layers are formed by randomly partitioning the input neurons and the corresponding weights of a network layer into two subsets (Left). For a specified output neuron of the layer, two internal features $y_1 = \mathbf{w}_1^{\top}\mathbf{x}_1$ and $y_2 = \mathbf{w}_2^{\top}\mathbf{x}_2$ can be computed from the partition (Right). Blank circles and dashed lines represent one sub-layer, and filled circles and solid lines represent the other. }  \label{FigIllus}
\end{figure}

In this section, we first use generic DNNs to present our proposed regularization method, and explain how our method is applied to their hidden layers. Our method is readily applied to fusion layers of deep networks that take practical multi-view data, which will be introduced in Section \ref{SecMultiViewRepLearning}.

We start with a DNN composed of $L$ FC layers. Denote its network parameters as $\Theta = \{ \mathbf{W}^l, \mathbf{b}^l \}_{l=1}^L$, where $\mathbf{W}^l$ and $\mathbf{b}^l$ are respectively the weight matrix and bias vector associated with the $l^{th}$ network layer. In the setting of supervised learning, given $M$ training samples ${\cal{S}} = \{\mathbf{s}^i\}_{i=1}^M$ of categorical data, the network parameters in $\Theta$ are optimized by minimizing the empirical risk $\frac{1}{M} \sum_{i=1}^M Loss(\mathbf{s}^i; \Theta )$, where $Loss(\cdot)$ is a properly chosen loss function, e.g., cross-entropy loss for image classification, and optimization is typically based on SGD or its variants \cite{Momentum}. As discussed in Section \ref{SecIntro}, DNNs of high model capacities are able to learn complex functions but susceptible to overfitting. To remedy, one may apply regularization to reduce their model capacities. Adding regularization to the network training objective results in the following optimization problem
\begin{equation}\label{EqnReguNNObj}
\min_{\Theta} \frac{1}{M} \sum_{i=1}^M Loss(\mathbf{s}^i; \Theta ) + \lambda R(\Theta) ,
\end{equation}
where $R(\cdot)$ is the regularizer to be specified, and $\lambda$ is a trade-off parameter.

In this work, we are interested in regularizing network training using CCA based multi-view learning criteria \cite{DCCA,DeepMultiViewRepLearn}. More specifically, for a specified $l^{th}$ network layer and a training sample $\mathbf{s}$, denote as $\mathbf{x} \in \mathbb{R}^{n_l}$ the feature vector of $\mathbf{s}$ computed all the way up from the input layer to the $l^{th}$ layer. The $l^{th}$ layer computes $f(\mathbf{z}) = f( \mathbf{W}^{\top}\mathbf{x} + \mathbf{b} ) \in \mathbb{R}^{n_{l+1}}$, where $f(\cdot)$ is an element-wise nonlinear activation function such as ReLU \cite{ReLU}, $\mathbf{W} \in \mathbb{R}^{n_l \times n_{l+1}}$ and $\mathbf{b} \in \mathbb{R}^{n_{l+1}}$ are the weight matrix and bias vector respectively, and where we omit the superscript $l$ to make the following notations of better clarity. By randomly partitioning $n_l$ input neurons/dimensions of the $l^{th}$ layer into two subsets \footnote{For simplicity, we only consider in this work the case of partitioning neurons of a hidden layer in a generic DNN into two subsets.}, we get feature subvectors $\mathbf{x}_1 \in \mathbb{R}^{n_{l_1}}$ and $\mathbf{x}_2 \in \mathbb{R}^{n_{l_2}}$ from $\mathbf{x}$, and the corresponding weight submatrices $\mathbf{W}_1 \in \mathbb{R}^{n_{l_1} \times n_{l+1}}$ and $\mathbf{W}_2 \in \mathbb{R}^{n_{l_2} \times n_{l+1}}$ from $\mathbf{W}$. We simply have $\mathbf{W}^{\top}\mathbf{x} = \mathbf{W}_1^{\top}\mathbf{x}_1 + \mathbf{W}_2^{\top}\mathbf{x}_2$. Discussions in Section \ref{SecIntro} suggest that $\mathbf{x}_1$ and $\mathbf{x}_2$ can be analogously considered as two feature views of the input data. The co-adaptation of features in each view is likely to cause network training to pay more attention to the extraction of view-specific patterns, rather than the category related patterns that are desired to be learned by network training. To address this issue and benefit more from both views of the features, one may use CCA criteria to regularize network training, which aim to increase the feature correlations between the two views. Given the two-view features $\mathbf{X}_1 = [\mathbf{x}_1^1, \dots, \mathbf{x}_1^M]$ and $\mathbf{X}_2 = [\mathbf{x}_2^1, \dots, \mathbf{x}_2^M]$ of training set ${\cal{S}}$, applying CCA to the $l^{th}$ network layer amounts to optimizing $\mathbf{W}_1$ and $\mathbf{W}_2$ by
\begin{eqnarray}\label{EqnCCA}
\max_{\mathbf{W}_1, \mathbf{W}_2} \frac{1}{M} \mathrm{tr}\left( \mathbf{W}_1^{\top} \mathbf{X}_1 \mathbf{X}_2^{\top} \mathbf{W}_2 \right) \quad\quad\quad\quad \\ \mathrm{s.t.} \ \frac{1}{M} \mathbf{W}_1^{\top} \mathbf{X}_1 \mathbf{X}_1^{\top}  \mathbf{W}_1  = \frac{1}{M} \mathbf{W}_2^{\top} \mathbf{X}_2 \mathbf{X}_2^{\top}  \mathbf{W}_2 = \mathbf{I} , \nonumber  
\end{eqnarray}
where $\mathbf{I}$ is an identity matrix of compatible size. The data matrices have been assumed centered for simplicity. Applying the above problem to the $l^{th}$ network layer also assumes implicitly that $n_{l+1} \leq \min(n_{l_1}, n_{l_2})$.

When directly using (\ref{EqnCCA}) as the regularizer $R(\cdot)$ in (\ref{EqnReguNNObj}), network training requires solving (\ref{EqnCCA}) with stochastic optimization methods. Unfortunately, as pointed out in \cite{StocOptmPCAPLS}, the objective (\ref{EqnCCA}) does not easily admit a stochastic optimization due to the involvement of data covariance matrices in the constraints. One may use batch gradient descent to solve (\ref{EqnCCA}). It computes gradients of the correlation objective w.r.t. the CCA projected features $\mathbf{W}_1^{\top} \mathbf{X}_1$ and $\mathbf{W}_2^{\top} \mathbf{X}_2$, which in turn will be used through back-propagation to compute the gradients w.r.t. $\mathbf{W}_1$ and $\mathbf{W}_2$, and w.r.t. all the network parameters in the layers below \cite{DCCA}. This is expensive as it involves computation of covariance matrices (of $\mathbf{W}_1^{\top} \mathbf{X}_1$ and $\mathbf{W}_2^{\top} \mathbf{X}_2$), their inverse square roots, and also performing matrix singular value decomposition (SVD). One may nevertheless try mini-batch based gradient descent to solve (\ref{EqnCCA}), which, however, may produce singular data covariance matrices; \cite{DCCA} also points out that solving (\ref{EqnCCA}) by mini-batch based stochastic optimization empirically gives unsatisfactory results. Given these challenges of directly using CCA as the regularizer $R(\cdot)$, we are motivated to find an alternative way to improve the correlations between the two feature views $\mathbf{X}_1$ and $\mathbf{X}_2$.

Inspired by batch normalization \cite{BatchNorm}, we propose to simplify the full CCA regularization in DNNs by considering the following two aspects. Firstly, instead of learning $\mathbf{W}_1$ and $\mathbf{W}_2$ to increase correlations at all the $n_{l+1}$ dimensions of the resulting features jointly, we propose to learn $\mathbf{w}_{1,i}$ and $\mathbf{w}_{2,i}$, $i \in \{ 1, \dots, n_{l+1} \}$, independently for each output neuron of the $l^{th}$ layer, where $\mathbf{w}_{1,i}$ and $\mathbf{w}_{2,i}$ are the $i^{th}$ columns of $\mathbf{W}_1$ and $\mathbf{W}_2$ respectively. Note that such a decoupling suggests that the resulting features at the $n_{l+1}$ dimensions could be correlated, but at the same time the constraint of $n_{l+1} \leq \min(n_{l_1}, n_{l_2})$ is also relaxed, enabling its flexible use in DNNs as tasks demand. Secondly, we use mini-batches of $m$ training samples, rather than all the $M$ ones, to approximate the statistics (i.e., mean and variance) necessary for computing correlations. This second simplification is enabled by the independent neuron-wise learning of $\mathbf{w}_{1,i}$ and $\mathbf{w}_{2,i}$: since in the joint case, the size $m$ of mini-batches is required to be big enough to avoid singularity of covariance matrices.

For a specified output neuron of the $l^{th}$ layer, we first introduce the (scalar) random variables $Y_1$ and $Y_2$ whose samples are respectively computed as $y_1 = \mathbf{w}_1^{\top}\mathbf{x}_1$ and $y_2 = \mathbf{w}_2^{\top}\mathbf{x}_2$, as illustrated in Figure \ref{FigIllus}, where we omit the neuron index for notational clarity. Given a mini-batch of $m$ training samples, we propose in this work the following {\it neuron-wise correlation-maximizing regularizer}
\begin{eqnarray}\label{EqnNeuronCorrRegu}
Corr(Y_1, Y_2) \approx \frac{ \sum_{i=1}^m (y_1^i - \mu_1)(y_2^i - \mu_2) }{ \sqrt{ \sigma_1^2\sigma_2^2 + \epsilon } } ,
\end{eqnarray}
where the scalar $\epsilon$ is introduced for numerical stability, and
\begin{eqnarray}
& y_1^i = \mathbf{w}_1^{\top}\mathbf{x}_1^i \ \ i = 1, \dots, m, &  \nonumber \\
& y_2^i = \mathbf{w}_2^{\top}\mathbf{x}_2^i \ \ i = 1, \dots, m, & \nonumber \\
& \mu_1 = \frac{1}{m}\sum_{i=1}^m y_1^i,     \ \ \mu_2 = \frac{1}{m}\sum_{i=1}^m y_2^i, & \nonumber \\
& \sigma_1^2 = \sum_{i=1}^m (y_1^i - \mu_1)^2, \ \ \sigma_2^2 = \sum_{i=1}^m (y_2^i - \mu_2)^2 . & \nonumber
\end{eqnarray}
Based on the neuron-wise regularizer (\ref{EqnNeuronCorrRegu}), we specify the general objective function (\ref{EqnReguNNObj}) as the following regularized problem to improve network training
\begin{eqnarray}\label{EqnNeuronCorrReguNNObj}
\min_{\Theta} \frac{1}{M} \sum_{i=1}^M Loss(\mathbf{s}^i; \Theta ) - \lambda \sum_{g \in {\cal{G}}} Corr({\cal{S}}; \theta_g) ,
\end{eqnarray}
where $g$ indexes the group ${\cal{G}}$ of neurons in the network layers that are specified to apply regularization, $\theta_g$ denotes a subset of network parameters $\Theta$ that are involved in the computation of the incoming features of the neuron $g$, and we have slightly abused the use of notation $Corr$ with that in (\ref{EqnNeuronCorrRegu}). Note that $\theta_g$ and $\theta_{g'}$ may contain overlapped parameters, i.e., those parameters associated with the common layers below $g$ and $g'$. The main objective (\ref{EqnNeuronCorrReguNNObj}) can be optimized using SGD or its variants \cite{Momentum}, by sampling mini-batches of training samples in iterative steps. Details are presented shortly. Complexity analysis presented in Section \ref{SecComplexityAnalysis} shows that compared with standard SGD training, our regularizer (\ref{EqnNeuronCorrRegu}) increases computation cost only by a constant factor.


\subsection{Correlation-Regularized Network Layer}

We still use the illustration in Figure \ref{FigIllus} as the running example. In the forward pass of a mini-batch of size $m$, any output neuron of the $l^{th}$ layer that is specified to apply the regularization (\ref{EqnNeuronCorrRegu}) computes $y_1^i = \mathbf{w}_1^{\top}\mathbf{x}_1^i$ and $y_2^i = \mathbf{w}_2^{\top}\mathbf{x}_2^i$, $i = 1, \dots, m$, which give output features of the neuron, before nonlinear activation $f(\cdot)$, as $z^i = y_1^i + y_2^i + b$, $i = 1, \dots, m$, where $b$ is the bias associated with this neuron. \footnote{For each neuron that is applied the regularization, we have ever introduced trainable scalar parameters $v_1$ and $v_2$ to re-scale the internal features $y_1$ and $y_2$, i.e., $z = v_1y_1 + v_2y_2 + b$, which is similar to the scheme introduced in \cite{WeightNorm}. In our experiments this alternative scheme does not necessarily improve the performance.} In the backward pass, we need to compute the gradients of the neuron-wise regularizer w.r.t. the weight vectors $\frac{\partial{Corr} }{ \partial{\mathbf{w}_1} }$ and $\frac{\partial{Corr} }{ \partial{\mathbf{w}_2} }$, and also those w.r.t. the input features $\frac{\partial{Corr} }{ \partial{\mathbf{x}_1^i} }$, $\frac{\partial{Corr} }{ \partial{\mathbf{x}_2^i} }$, $i = 1, \dots, m$. These gradients can be derived via multi-variable chain rule, and we give their explicit forms in Appendix \ref{AppendixCorrRegGradients}. Note that the later ones are used to compute through back-propagation gradients of the regularizer w.r.t. network parameters in the lower layers that are also involved in the computation of $\{ z^i \}_{i=1}^m$.

We usually apply (\ref{EqnNeuronCorrRegu}) to all the $n_{l+1}$ output neurons of the specified $l^{th}$ layer. To make regularized network training efficient, we note that for these $n_{l+1}$ neurons, their respective internal features and statistics, i.e., $\{ y_1^i \}_{i=1}^m$, $\{ y_2^i \}_{i=1}^m$, $\mu_1$, $\mu_2$, $\sigma_1^2$, and $\sigma_2^2$, and also the corresponding gradients can be computed independently and in parallel. Given the mini-batch of layer inputs of the two-view features $\mathbf{x}^i = [ \mathbf{x}_1^i, \mathbf{x}_2^i ]$, $i = 1, \dots, m$, we write gradients of the $n_{l+1}$ neuron-wise regularizers in the compact forms as $\left[ \frac{\partial{\mathbf{Corr}} }{ \partial{\mathbf{x}_1^i} } , \frac{\partial{\mathbf{Corr}} }{ \partial{\mathbf{x}_2^i} } \right]$, $i = 1, \dots, m$, and $\left[ \frac{\partial{\mathbf{Corr}} }{ \partial{\mathbf{W}_1} } , \frac{\partial{\mathbf{Corr}} }{ \partial{\mathbf{W}_2} } \right]$, where $\mathbf{Corr}$ denotes the $n_{l+1}$ correlation objectives compactly. More specifically, $\frac{\partial{\mathbf{Corr}} }{ \partial{\mathbf{x}_1^i} }$ \emph{sums} the gradients of neuron-wise regularizers in the layer w.r.t. the input $\mathbf{x}_1^i$, and $\frac{\partial{\mathbf{Corr}} }{ \partial{\mathbf{W}_1} }$ \emph{independently computes} the gradient of each neuron-wise regularizer w.r.t. its associated weight vector $\mathbf{w}_1$; the same operations apply to $\frac{\partial{\mathbf{Corr}} }{ \partial{\mathbf{x}_2^i} }$ and $\frac{\partial{\mathbf{Corr}} }{ \partial{\mathbf{W}_2} }$.

In other words, we implement our proposed scheme (\ref{EqnNeuronCorrRegu}) as \emph{a correlation-regularized network layer} (CorrReg): in the forward pass, the computation is the same as a standard network layer, i.e., we do not explicitly compute the internal features $\mathbf{y}_1 = \mathbf{W}_1^{\top}\mathbf{x}_1$ and $\mathbf{y}_2 = \mathbf{W}_2^{\top}\mathbf{x}_2$, and instead we directly compute $\mathbf{z} = \mathbf{W}^{\top}\mathbf{x} + \mathbf{b}$, followed by element-wise nonlinear activation; in the backward pass, we compute the gradients of the correlation-regularized loss w.r.t. layer weights and layer inputs, which we spell out as
\begin{eqnarray}
& \frac{1}{m} \sum_{i=1}^m \frac{ \partial{Loss} (\mathbf{x}^i) }{ \partial{\mathbf{W}} }  - \lambda \left[ \frac{\partial{\mathbf{Corr}} }{ \partial{\mathbf{W}_1} }, \frac{\partial{\mathbf{Corr}} }{ \partial{\mathbf{W}_2} } \right] & \nonumber \\
& \frac{ \partial{Loss} (\mathbf{x}^i) }{ \partial{\mathbf{x}^i} }  - \lambda \left[ \frac{\partial{\mathbf{Corr}} }{ \partial{\mathbf{x}_1^i} }, \frac{\partial{\mathbf{Corr}} }{ \partial{\mathbf{x}_2^i} } \right]  \ i = 1, \dots, m , &  \nonumber
\end{eqnarray}
where $\frac{ \partial{Loss} (\mathbf{x}^i) }{ \partial{\mathbf{W}} }$ and $\frac{ \partial{Loss} (\mathbf{x}^i) }{ \partial{\mathbf{x}^i} }$ can be computed via standard back-propagation. The gradient of the main loss w.r.t. the layer bias vector $\mathbf{b}$ can be obtained in the same way.


\vspace{0.1cm}
\subsubsection{Analysis of Computational Complexity}
\label{SecComplexityAnalysis}

We analyze the additional computation cost incurred by imposing CorrReg on a network layer. Consider an $l^{th}$ layer that computes $f(\mathbf{z}) = f( \mathbf{W}^{\top}\mathbf{x} + \mathbf{b} ) \in \mathbb{R}^{n_{l+1}}$ for a mini-batch of $m$ samples, and that has layer parameters $\mathbf{W} \in \mathbb{R}^{n_l \times n_{l+1}}$ and $\mathbf{b} \in \mathbb{R}^{n_{l+1}}$. Assume arithmetic with individual elements has complexity ${\cal{O}}(1)$. In the forward pass, the computation for the layer with or without CorrReg is the same, and its complexity is ${\cal{O}}(mn_ln_{l+1})$. In the backward pass, without using CorrReg the complexity for back-propagating gradients through this layer is ${\cal{O}}(mn_ln_{l+1})$. By simplifying the gradient formulas given in Appendix \ref{AppendixCorrRegGradients} and also writing them in matrix forms, we have the same complexity of ${\cal{O}}(mn_ln_{l+1})$ when using CorrReg. In summary, the complexity by imposing CorrReg on a network layer increases only by a constant factor. In contrast, using the CCA objective (\ref{EqnCCA}) as the regularizer involves computing inverse square root of covariance matrices of the size $n_{l+1}\times n_{l+1}$, and also performing SVD for matrix of the same size (one may refer to \cite{DCCA} for gradient formulas of CCA objective); it has the overall complexity of ${\cal{O}}( mn_ln_{l+1} + mn_{l+1}^2 + n_{l+1}^3 )$ in the backward pass, which is significantly worse than that of CorrReg.

\subsection{Correlation-Regularized Convolutional Networks}
\label{SecCorrRegConvNets}

Our proposed CorrReg can regularize both FC and conv layers. As discussed above, for an FC layer computing $f(\mathbf{z}) = f( \mathbf{W}^{\top}\mathbf{x} + \mathbf{b} )$, we apply regularization to the input $\mathbf{z}$ of $f$ by performing a random two-way partition on feature dimensions of $\mathbf{x}$, producing two internal features $\mathbf{y}_1 = \mathbf{W}_1^{\top}\mathbf{x}_1$ and $\mathbf{y}_2 = \mathbf{W}_2^{\top}\mathbf{x}_2$, where the partition is fixed once determined. CorrReg is indeed to improve correlations between each dimension pair of $\mathbf{y}_1$ and $\mathbf{y}_2$. It is straightforward to extend the above scheme of \emph{single} two-way partition in CorrReg to its version of \emph{multiple} two-way partitions. For a specified FC layer, one may simply perform multiple, random two-way partitions on feature dimensions of $\mathbf{x}$; each of them produces their respective internal features, and also their respective gradients w.r.t. layer weights and layer inputs. The overall regularization imposed on this layer can be obtained by averaging the gradients from these multiple two-way partitions. Experiments investigating the efficacy of this scheme of multiple two-way partitions are reported in Section \ref{SecExpImageClassification}.

For a conv layer, we perform single or multiple random two-way partitions on its input feature maps in the same way as for an FC layer. Each two-way partition produces two internal feature maps (corresponding to $y_1$ and $y_2$ in Figure \ref{FigIllus}) for each output feature map of the layer. Although features/observations at nearby locations of an image are generally correlated, we do not explicitly exploit such correlations. Instead, we independently apply regularization at each spatial location/pixel of each output feature map of the layer, so that correlations between the corresponding spatial locations in each pair of the internal feature maps are improved, where regularization is again applied before nonlinear activation. Applying our proposed CorrReg to modern architectures of DNNs (e.g., ConvNets \cite{VGGNet}, variants of ResNets \cite{ResNet,WideResNet}, or DenseNets \cite{DenseNet}) is very simple: one simply replaces FC or conv layers with their CorrReg versions.


\section{Use of CorrReg Fusion in Deep Representation Learning from Multi-View Data}
\label{SecMultiViewRepLearning}

\begin{figure}[t]
\centering

\includegraphics[scale=0.45]{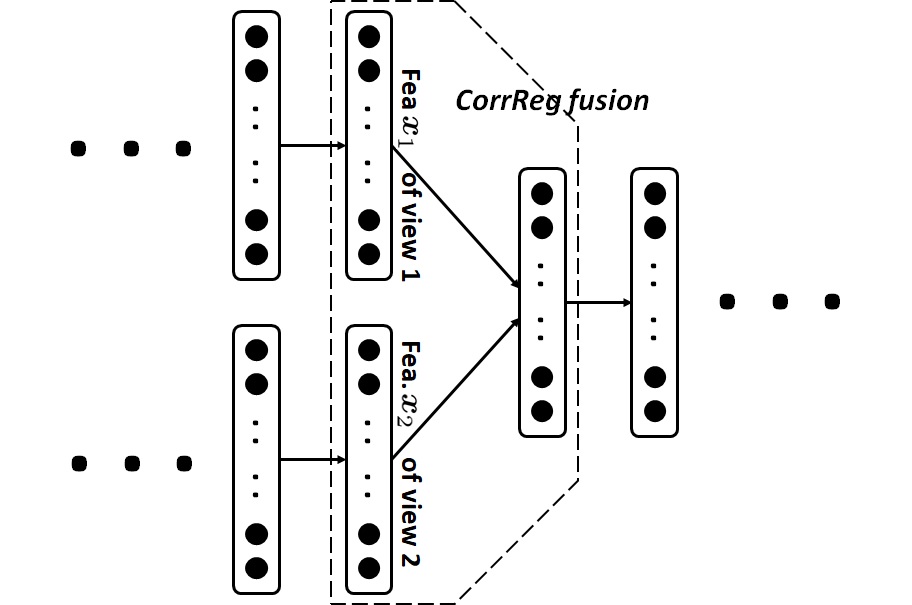}  \\

\caption{Illustration of a \emph{CorrReg fusion} layer (inside the dashed box).   }  \label{FigCorrRegFusionIllus}
\end{figure}

We present in this section how CorrReg can be readily applied to deep networks that take as inputs data of multiple views/modalities and learn deep representations from them. Different from generic DNNs, these networks by design have lower, parallel streams for data of individual views, and it is natural to apply CorrReg to the \emph{fusion} layers where features of different views are concatenated for use in the subsequent layers. The fusion layer in such a network is usually based on FC or conv layers. To use CorrReg, one may simply treat output features of two parallel streams as the input two-view features of CorrReg (corresponding to $\mathbf{x}_1$ and $\mathbf{x}_2$ in Section \ref{SecCorrReg}), and replace the fusion layer with its CorrReg version, resulting in a \emph{CorrReg fusion} layer. Figure \ref{FigCorrRegFusionIllus} gives the illustration. In the following, we take network architectures used in our RGB-D recognition experiments to instantiate the use of CorrReg fusion.

Our networks for RGB-D object recognition and scene recognition are based on ConvNets \cite{AlexNet} and ResNets \cite{ResNet}. Take an 8-layer ConvNet as the example. We modify it by enabling it to take inputs of both RGB and depth channels. The 8-layer ConvNet is composed of two lower, parallel streams followed by an upper, single stream. Outputs of the two lower streams are concatenated as inputs of the upper stream, and CorrReg is applied to the first layer of the upper stream, which thus becomes a CorrReg fusion layer. In this work, we also investigate the effects when the CorrReg fusion layer is at different ``heights'' (lower, middle, or upper layers) of the network. Figure \ref{FigRGBDRecogAlexNet} gives the illustration. Adaptation of ResNets is similar as above. We use such adapted network architectures for experiments of RGB-D object recognition and scene recognition in Section \ref{SecExpRGBDRecog}.

\section{Relations with Existing Works}
\label{SecLiterature}

Our proposed CorrReg method is related to four categories of existing research: regularization of generic DNNs, deep learning research that specially focuses on multi-view representation learning, RGB-D object/scene recognition, and multi-view recognition of 3D object shapes. We respectively discuss the relations as follows.

\vspace{0.1cm}
\noindent\textbf{Network regularization} In the literature of DNNs, various regularization techniques have been proposed to address the issue of overfitting in network training, including the traditional early stopping,  weight decay, and data augmentation \cite{MultiColumnCNN}, and also the more recent Dropout \cite{Dropout}, Dropconnect \cite{DropConnect}, batch normalization \cite{BatchNorm}, and all conv layer based networks \cite{NIN,AllCNN}. Among them, Dropout and Dropconnect are most related to our proposed method. In the original proposal of Dropout \cite{Dropout}, each hidden neuron is randomly dropped (usually with a probability of $0.5$) at each training iteration, and the network is then updated on weights that are connected to the remaining neurons. During inference, all network weights are used after halving their values. Baldi and Sadowski \cite{UnderstandDropout} quantitatively analyze that random operations of dropout training and the associated inference can be understood as a good approximation to the expectation of outputs of a subnetwork ensemble, by introducing a bridging quantity Normalized Weighted Geometric Mean. They further show that the expectation of dropout gradients w.r.t. a network weight is approximately the gradient of the subnetwork ensemble regularized by {\it adaptive weight decay}.  Wager {\it et al.} \cite{DropoutAdapRegu} present alternative interpretation of dropout training as adaptive weight decay by treating dropout as feature noising in generalized linear models. Analysis similar to \cite{UnderstandDropout} can be applied to Dropconnect \cite{DropConnect} by randomly dropping weight connections rather than network neurons.

Dropout and Dropconnect achieve regularization by first sampling features/subnetworks (of shared weights), and then averaging over outputs of the subnetwork ensemble. Different from them, our CorrReg scheme explicitly increases the correlations between (internal) features of different views, and regularization is achieved by suppressing view-specific noisy patterns. We empirically show in this work the usefulness of CorrReg in improving network training, and leave its theoretical connections with classical regularization to future research. Note that a few recent methods of network regularization explicitly reduce correlations across dimensions of the output features of a layer \cite{DeCov}, or achieve similar effects by enforcing orthogonality of the layer weight matrix \cite{LocallyConstrainedDeCorr,SVB}. These methods impose regularization complementary to our proposed CorrReg, and we are interested in the investigation of their combined use in future research.

\vspace{0.1cm}
\noindent\textbf{Deep multi-view representation learning} Recent deep multi-view representation learning methods include those based on CCA \cite{DCCA,DeepMultiViewRepLearn,CorrNet} and those based on auto-encoders (AE) \cite{MultimodalDL}. AE based methods typically learn a shared bottleneck layer on top of lower view-specific layers, and the learned joint representation at the bottleneck layer is used for reconstruction of multiple views. Deep CCA \cite{DCCA} directly applies CCA to the output layers of two deep networks, so that the learned networks can produce maximally correlated features at the output layers. Wang {\it et al.} \cite{DeepMultiViewRepLearn} extend deep CCA as Deep Canonically Correlated Auto-Encoders, by balancing the correlation objective between the two views with their respective reconstruction objectives.

Most of existing deep multi-view learning works take a two-stage strategy for the final tasks of interest: they first learn from data of multiple views/modalities deep features in a common space, and then use the learned features in the common space either to train classifiers for multi-view or across-view classification, or to reconstruct data of missing views. In contrast, our use of correlation based multi-view learning is to regularize training of end-to-end networks, where parameters that project multi-view data into a common space are exactly those of an intermediate network layer.

\vspace{0.1cm}
\noindent\textbf{RGB-D object and scene recognition} RGB-D object recognition \cite{HMP4RGBDRecog,eitelmultimodal,MultiModalDBM,WangMMSS} has drawn research attention recently as a typical application of multi-view learning. Lai \emph{et al.} \cite{RGBDObjDataset} collect the first large-scale, hierarchial RGB-D object dataset using a Kinect camera; they show that depth information substantially helps object recognition, by concatenating hand-crafted depth features (e.g., spin images \cite{SpinImage}) with those of RGB ones, and using the concatenated features for classification. In \cite{ConvRecursive3DObjRecog}, a hierarchical learning model of Convolutional and Recursive Neural Networks (CNNs and RNNs) are proposed, where CNNs are used for learning low-level features and RNNs with random weights for efficiently extracting higher-order features; this combined model is applied to RGB and depth images separately, and the resulting features are concatenated for classification.

Eitel \emph{et al.} \cite{eitelmultimodal} propose a multi-modal deep learning architecture for RGB-D object recognition, which fuses, via feature concatenation, outputs of two parallel streams of modality-specific subnetworks (composed of conv and FC layers), and uses two additional FC layers for feature transformation and softmax classification; the whole network is trained via standard back-propagation with no consideration of multi-view learning/regularization criteria. Built on top of two parallel streams of ResNets (after removing their respective last layers of classifier) \cite{ResNet}, a Correlated and Individual Multi-modal (CIM) learning layer is proposed in \cite{CIM4MultiModalDL}; CIM aims to learn, in a discriminative and complementary manner, both correlated and modality-specific features from output vectors of the two ResNets, where ``correlation'' is measured by the Euclidean distance between projected features of the two ResNets' outputs; parameters of the whole network in \cite{CIM4MultiModalDL} are updated in an alternating manner: those of the two lower streams of ResNets are updated after updating of the CIM parameters (projection matrices) converges. In \cite{MDSICNN}, a deep learning framework termed MDSI-CNN is proposed to learn highly discriminative and spatially invariant multi-modal feature representations at different hierarchical levels, which is technically achieved by introducing spatial transformer network \cite{STN} and Fisher encoding into CNN architectures. The problem of RGB-D image classification with limited training samples is addressed in \cite{LCCRRD}. It takes a domain adaptation approach and enforces the prediction consistency between two classifiers that are respectively learned either from the combined RGB and depth features or from the RGB features alone. Results on RGB-D object recognition show the efficacy of the proposed approach.

Methods for RGB-D scene recognition \cite{SSCNN,FVCNN,song2017depth,DBSNN} largely follow those of RGB-D object recognition. In particular, multi-modal deep architectures similar to that of \cite{eitelmultimodal} are still the main workhorse to get good recognition performance. We also use such a type of networks for RGB-D recognition, but with our proposed CorrReg fusion layers that have in-built neuron-wise correlation regularization. Training of CorrReg fusion based networks has no difference from standard back-propagation.

\vspace{0.1cm}
\noindent\textbf{Multi-view recognition of 3D object shapes} Recent research shows that multi-view images are of a promising representation for recognition of 3D object shapes. Given a 3D object model (mesh), multiple 2D images can be rendered by placing virtual cameras around the object, and recognition is based on the rendered 2D images of multiple views. Among recent methods, MVCNN \cite{su2015multi} is a representative one that uses parallel streams of conv layers to extract features from individual views, and then aggregates these features as a global signature simply via max pooling across different views. Subsequent works improve over MVCNN by strengthening interaction among feature learning of individual views. For example, MHBN \cite{yu2018multi} uses harmonized bilinear pooling to aggregate local features, and GVCNN \cite{feng2018gvcnn} proposes a group-view framework to model correlations among different views at a hierarchy of multiple levels. In this work, we adapt the architectural design of MVCNN by incorporating into it CorrReg fusion layers. We use such an adapted architecture to verify the usefulness of CorrReg for multi-view based 3D object recognition.

\section{Experiments}
\label{SecExp}

\begin{table*}[ht]
\caption{ Error rates ($\%$) on CIFAR-10 \cite{Cifar} when applying CorrReg, with varying numbers $\texttt{nReg}$ of random two-way partitions, to different layers of a variant of LeNet \cite{LeCun98}. Setting $\texttt{nReg}$ as $0$ indicates no CorrReg is applied to any network layer. Experiments of each setting are run for $5$ times, and results are in the format of mean (standard deviation).   }  \label{TabExpLeNet}
\begin{center}
\begin{tabular}{|l|c|c|c|c|c|c|}
\hline                      & No CorrReg & CorrReg Conv2 & CorrReg Conv3 & CorrReg FC4 & CorrReg FC5 & CorrReg FC6  \\
\hline $\texttt{nReg} = 0$  & $17.42$ ($0.16$) & - & - & - & - & -    \\
\hline $\texttt{nReg} = 1$  &    -             & $17.15$ ($0.11$) & $17.14$ ($0.25$) & $17.13$ ($0.27$) & $16.75$ ($0.20$) & $16.92$ ($0.30$)    \\
\hline $\texttt{nReg} = 3$  &    -             & $17.17$ ($0.21$) & $17.19$ ($0.17$) & $17.25$ ($0.23$) & $16.81$ ($0.28$) & $17.08$ ($0.11$)    \\
\hline $\texttt{nReg} = 5$  &    -             & $17.08$ ($0.13$) & $17.23$ ($0.23$) & $17.28$ ($0.15$) & $16.64$ ($0.20$) & $16.84$ ($0.18$)    \\
\hline $\texttt{nReg} = 10$ &    -             & $17.15$ ($0.40$) & $17.12$ ($0.12$) & $17.16$ ($0.20$) & $16.87$ ($0.18$) & $17.14$ ($0.31$)    \\
\hline
\end{tabular}
\end{center}
\end{table*}

In this section, we first present control experiments of image classification to investigate the effectiveness of our proposed CorrReg for regularization of network training. We use generic DNNs including ConvNet (LeNet) \cite{LeCun98}, and modern deep architectures of ResNet \cite{ResNet}, Wide ResNet \cite{WideResNet}, DenseNet \cite{DenseNet}, and ResNeXt \cite{ResNeXt}. These experiments are conducted on the benchmark datasets of CIFAR-10, CIFAR-100 \cite{Cifar}, and ImageNet \cite{ILSVRC15}. We then present experiments of RGB-D object/scene recognition and multi-view 3D object recognition to evaluate the usefulness of CorrReg for practical multi-view learning problems. We use the benchmark datasets of RGB-D object \cite{RGBDObjDataset}, RGB-D scene \cite{SUNdataset}, and ModelNet40 \cite{wu20153d} for these experiments, and compare with the state-of-the-art results.

We use cross-entropy loss to train all these networks. Training is based on SGD with momentum. Without mentioning otherwise, we use mini-batches of size $128$, momentum of $0.9$, and weight decay of $0.0001$; network parameters are initialized using Gaussian random weights; batch normalization is applied, before ReLU nonlinearity, in all networks to accelerate their training. In each experiment, the initial learning rate, the value of $\lambda$ in (\ref{EqnNeuronCorrReguNNObj}), and also the dropping rate of Dropout (when using Dropout regularization) are determined by using $10\%$ of training samples as the validation set. As (\ref{EqnNeuronCorrReguNNObj}) suggests, we use constant $\lambda$ values for all neurons that are specified to apply CorrReg. Learning rates are decayed at the rate of $0.1$ when learning curves plateau. Our implementation and experiments are based on the Torch library \cite{torch}.

\subsection{Control Experiments of Image Classification}
\label{SecExpImageClassification}

We use the CIFAR-10 dataset for our controlled studies on a plain ConvNet (a variant of LeNet \cite{LeCun98}). The CIFAR-10 dataset consists of $10$ object categories of $60,000$ color images of size $32\times 32$ ($50,000$ training and $10,000$ testing ones). We follow \cite{MaxoutNetwork} and preprocess the data using global contrast normalization and ZCA whitening. Our used LeNet variant consists of $3$ conv layers (the first one is the input layer), followed by $3$ FC layers (the last one is the output layer). Max or average pooling layers are applied after each conv layer. More layer specifics are given in Appendix \ref{AppendixNetArchitecture}.

We first investigate the regularization effects when applying CorrReg to different network layers. To this end, we replace each of the network layers (except the input one), namely Conv2, Conv3, FC4, FC5, and FC6, with their CorrReg versions respectively, and compare the recognition performance. As indicated in Section \ref{SecCorrRegConvNets}, the scheme of multiple (random) two-way partitions can be used when applying CorrReg to any network layer. We also investigate how different numbers $\texttt{nReg}$ of two-way partitions in CorrReg achieve regularization, for which we set $\texttt{nReg} = 1, 3, 5$, or $10$. Note that when $\texttt{nReg} = 1$, we use the first half neurons/feature maps of the layer as a subset, and use the other half as the second subset; when $\texttt{nReg} > 1$, regularization is achieved by averaging over those of the multiple two-way partitions. We run experiments of each setting (the layer/$\texttt{nReg}$ pair) for $5$ times, and report results in the format of mean (standard deviation). Error rates reported in Table \ref{TabExpLeNet} tell that applying CorrReg, with any number $\texttt{nReg}$ of two-way partitions, to these layers consistently achieves performance boost over the LeNet variant baseline. In general, CorrReg is more effective for (upper) FC layers; this is reasonable since a densely connected FC layer contains much more trainable parameters than a conv layer does, and is thus more susceptible to overfitting. Setting $\texttt{nReg} > 1$ sometimes helps in getting even better results, but at the cost of slightly increased computation. In the subsequent experiments, we simply set $\texttt{nReg} = 1$ for computational efficiency.

As a technique for regularization of network training, CorrReg is related to the methods \cite{Dropout,DropConnect,BatchNorm} discussed in Section \ref{SecLiterature}, in particular Dropout \cite{Dropout}. To compare CorrReg with Dropout, we apply them to the FC5 layer of LeNet variant. Since their working mechanisms are different, one might be also interested in using them together. Table \ref{TabExpDropoutComp} reports the comparative results. CorrReg achieves improvement comparable to that of Dropout, where the dropping rate is optimally set as $0.2$ by tuning from the range of $(0, 1)$ on the validation set. Using CorrReg together with Dropout further improves the performance, showing the complementary regularization benefit of CorrReg to that of Dropout.

\begin{table}[h]
\caption{Comparison of image classification on CIFAR-10 \cite{Cifar} when applying CorrReg and/or Dropout to an upper FC layer of a variant of LeNet \cite{LeCun98}. Experiments are run for $5$ times, and results are in the format of mean (standard deviation).  }  \label{TabExpDropoutComp}
\begin{center}
\begin{tabular}{cc}
\hline Methods                & Error rates ($\%$)  \\
\hline {\scriptsize Plain LeNet variant}    & $17.42$ ($0.16$) \\
       {\scriptsize Dropout \cite{Dropout}} & $16.72$ ($0.18$) \\
       {\scriptsize CorrReg}                & $16.75$ ($0.20$) \\
       {\scriptsize CorrReg + Dropout}      & $16.36$ ($0.25$) \\
\hline
\end{tabular}
\end{center}
\end{table}

CorrReg achieves regularization via improving correlations between the internal features produced by two-way partition of a layer, which suggests a natural alternative that halves the number of layer neurons. Halving the number of layer neurons reduces the model capacity and creates ``bottlenecking'' of information flow, thus implicitly imposing regularization. Oppositely, one may be also interested in alternatives that increase the number of layer neurons with varying factors. To investigate the efficacy of CorrReg for models with different capacities, we apply these alternatives again to the FC5 layer of LeNet variant. Results in Table \ref{TabExpModelCapacityComp} show that larger models perform better than smaller ones do, and applying our proposed CorrReg to larger models further improves the performance. We also compute in Table \ref{TabExpModelCapacityComp} the averaged (pair-wise) correlation among features of training samples learned at different layer neurons, in order to understand how network capacities relate to the behavior of feature correlations across layer neurons and how CorrReg plays a role here as a regularization. Results show that as the numbers of layer neurons increase, feature correlations between layer neurons increase, and applying CorrReg further enhances this effect.

\begin{table}[h]
\caption{Experiments on CIFAR-10 \cite{Cifar} using a variant of LeNet \cite{LeCun98}. CorrReg is optionally applied to an upper FC layer with varying numbers of layer neurons. Results are in the format of error rate ($\%$)/correlation coefficient ($1\mathrm{e}^{-2}$). Refer to the main text for how correlation is computed. }  \label{TabExpModelCapacityComp}
\begin{center}
\begin{tabular}{ccccc}
\hline {\scriptsize Neuron No.}     & 128 & 256 & 512 & 1024   \\
\hline {\scriptsize W/O CorrReg}    & $17.85$/$0.53$ & $17.42$/$0.70$ & $17.17$/$0.76$ & $17.01$/$0.79$ \\
\hline {\scriptsize With CorrReg}   & $17.35$/$0.65$ & $16.75$/$0.75$ & $16.35$/$0.79$ & $16.08$/$0.82$ \\
\hline
\end{tabular}
\end{center}
\end{table}

CorrReg is a neuron-wise scheme of CCA regularization. One might be interested in the performance when using CCA as the regularizer. To this end, we apply the CCA objective (\ref{EqnCCA}) as a regularizer to the FC5 layer of the LeNet variant, where regularization parameter is set as $1\mathrm{e}^{-6}$ by optimally tuning on the validation set. Computational complexity of CCA regularization is significantly worse than that of CorrReg, and Table \ref{TabExpCCAComp} shows that it practically consumes more time per iteration of SGD training (measured on an M40 GPU and Intel Xeon CPU running at 2.2 GHz). Although CCA regularization improves performance over that of plain LeNet variant, its results with different sizes of mini-batches are worse than those of CorrReg; we hypothesize that this is because optimization of CCA objective (particularly the constraints in (\ref{EqnCCA})) is incompatible with SGD based network training.

\begin{table*}[ht]
\caption{Computation and recognition performance on CIFAR-10 \cite{Cifar} when applying CorrReg or CCA regularization, with different sizes $m$ of mini-batches, to an upper FC layer of a variant of LeNet \cite{LeCun98}. Wall-clock time is measured on an M40 GPU and Intel Xeon CPU running at 2.2 GHz. Experiments are run for $5$ times, and accuracies are in the format of mean (standard deviation). $n_l = 256$ and $n_{l+1} = 64$ denote the numbers of input and output neurons of the layer respectively. }  \label{TabExpCCAComp}
\begin{center}
\begin{tabular}{|c|ccc|ccc|ccc|}
\hline Methods                                     & \multicolumn{3}{c|}{Plain LeNet variant} &  \multicolumn{3}{c|}{CorrReg} & \multicolumn{3}{c|}{CCA regularization}       \\
                                                   & $m=128$ & $m=256$ & $m=512$ & $m=128$  & $m=256$ & $m=512$ & $m=128$ & $m=256$ & $m=512$      \\
\hline {\scriptsize Computational}      &  &&  &  &&  &  &&        \\
       {\scriptsize Complexity}      &  \multicolumn{3}{c|}{{\scriptsize ${\cal{O}}(mn_ln_{l+1})$}}   &   \multicolumn{3}{c|}{{\scriptsize ${\cal{O}}(mn_ln_{l+1})$}}  & \multicolumn{3}{c|}{{\scriptsize ${\cal{O}}(mn_ln_{l+1} + mn_{l+1}^2 + n_{l+1}^3)$}}        \\
\hline {\scriptsize wall-clock time}   &  &&  &  &&  &  &&        \\
       {\scriptsize per iter. (sec.)}      & {\scriptsize $0.009$}  & {\scriptsize $0.016$}  & {\scriptsize $0.030$}  &  {\scriptsize $0.013$}  & {\scriptsize $0.020$}  & {\scriptsize $0.035$}  & {\scriptsize $0.025$}  & {\scriptsize $0.035$}  &  {\scriptsize $0.048$}      \\
\hline {\scriptsize Error rates ($\%$)}            & {\scriptsize $17.42$ ($0.16$)}  & {\scriptsize $17.64$ ($0.17$)}  & {\scriptsize $18.33$ ($0.19$)}  &  {\scriptsize $16.75$ ($0.20$)}  & {\scriptsize $17.13$ ($0.18$)}  & {\scriptsize $17.43$ ($0.12$)}  & {\scriptsize $17.24$ ($0.33$)}  & {\scriptsize $17.39$ ($0.22$)}  & {\scriptsize $17.65$ ($0.24$)}      \\
\hline

\end{tabular}
\end{center}
\end{table*}

\begin{figure}[t]
\centering

\includegraphics[scale=0.18]{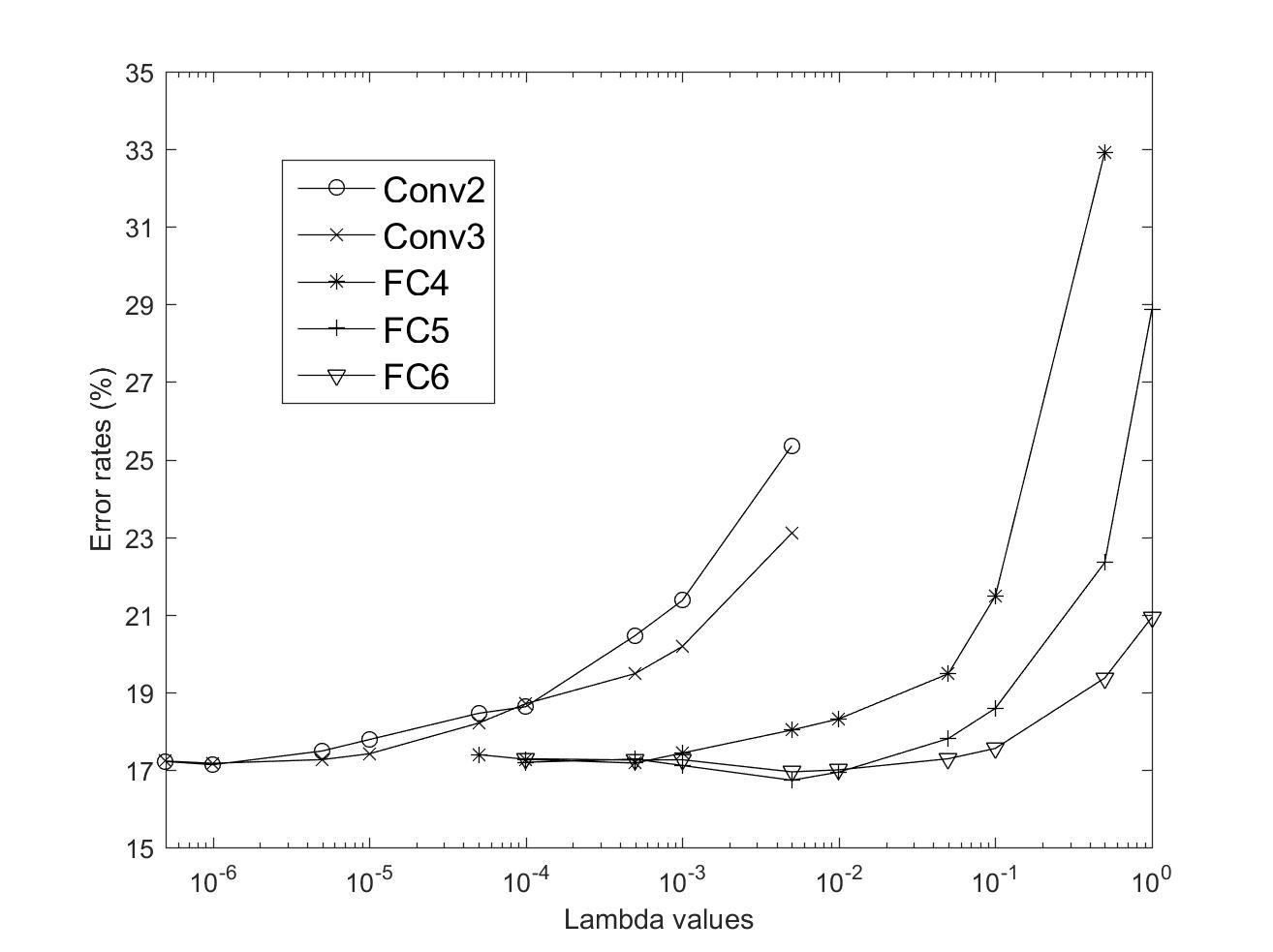}   \\

\caption{Investigation of the penalty parameter $\lambda$ on CorrReg's performance. Each line represents the validation errors on CIFAR-10 \cite{Cifar} when applying CorrReg to a layer of a variant of LeNet \cite{LeCun98} using a range of $\lambda$ values. Each error rate is a mean out of $5$ runs. These results are obtained by using $10\%$ of CIFAR-10 training samples as validation. }  \label{FigLambdaTuning}
\end{figure}

The penalty parameter $\lambda$ in (\ref{EqnNeuronCorrReguNNObj}) controls the amount of regularization that CorrReg imposes on network training. To investigate how performance of CorrReg depends on $\lambda$ values, we use $10\%$ of training samples as validation, and apply CorrReg to different layers of the LeNet variant (i.e., the settings of Table \ref{TabExpLeNet} with $\texttt{nReg} = 1$) using a range of $\lambda$ values. Results are plotted in Figure \ref{FigLambdaTuning}. Figure \ref{FigLambdaTuning} shows that smaller values of $\lambda$ are usually better when applying CorrReg to (lower) conv layers, and larger values of $\lambda$ are usually better for (upper) FC layers. This is reasonable due to two compound reasons: (1) when applying CorrReg to an upper network layer, larger values of $\lambda$ are needed in order to back-propagate the regularization for better learning of features/parameters of all the layers below; (2) conv layers already have intrinsic regularization via weight sharing. This inconsistency of optimal $\lambda$ values across different network layers makes use of CorrReg less convenient. Fortunately, results in this section suggest that to get the most effective regularization, one may simply apply CorrReg to an upper (FC) network layer, and set the optimal $\lambda$ values accordingly. Setting $\lambda \in [1\mathrm{e}^{-3}, 1\mathrm{e}^{-1}]$ typically gives good results. Experiments in the subsequent sections follow this empirical rule.

\vspace{0.1cm}
\subsubsection{Results on Modern Deep Architectures}
\label{SecExpModernNetImgClassification}

\begin{table*}[ht]
\caption{Results (error rates $\%$) on CIFAR-10 and CIFAR-100 \cite{Cifar} of several deep architectures with or without regularization. Standard data augmentation is used as in \cite{DeeplySupervisedNet}. Refer to the main text for how Dropout-S1, Dropout-S1S2, CorrReg, and CorrReg-DropoutS2 are applied to these architectures. } \label{TabExpModernNetImgClassification}
\begin{center}
\begin{tabular}{|c|ccccc|ccccc|}
\hline   {\scriptsize Architecture}  &                         &                          & {\scriptsize CIFAR-10}     &                       &                                 &                         &                  & {\scriptsize CIFAR-100}                          &                            &  \\
                                     & {\scriptsize W/O Regu.} & {\scriptsize Dropout-S1} & {\scriptsize Dropout-S1S2} & {\scriptsize CorrReg} & {\scriptsize CorrReg-}  & {\scriptsize W/O Regu.} & {\scriptsize Dropout-S1} & {\scriptsize Dropout-S1S2} & {\scriptsize CorrReg} & {\scriptsize CorrReg-}  \\
                                     &                         &                          &                            &                       & {\scriptsize DropoutS2} &                         &                          &  &  & {\scriptsize DropoutS2}  \\
\hline {\scriptsize ResNet \cite{PreActResNet}}      & $6.69$      & $6.25$    & $6.13$ & $6.15$ & $\mathbf{6.02}$ & $27.68$ & $28.08$ & $26.58$ & $27.32$ & $\mathbf{26.33}$  \\

\hline {\scriptsize Wide ResNet \cite{WideResNet}}   & $4.39$      & $4.50$    & $4.22$ & $4.11$ & $\mathbf{4.05}$ & $21.40$ & $22.07$ & $20.75$ & $21.32$ & $\mathbf{20.38}$\\

\hline {\scriptsize DenseNet \cite{DenseNet}}        & $3.90$      & $3.81$    & $3.75$ & $\mathbf{3.45}$ & $3.51$ & $18.99$ & $18.19$ & $18.15$ & $18.06$ & $\mathbf{17.99}$ \\

\hline
\end{tabular}
\end{center}
\end{table*}

We further investigate the regularization effects of CorrReg on modern deep architectures. For the datasets of CIFAR-10 and CIFAR-100, we use the representative architectures of ResNet \cite{PreActResNet}, Wide ResNet \cite{WideResNet}, and DenseNet \cite{DenseNet}. The CIFAR-100 dataset is an adaptation of CIFAR-10, consisting of $100$ object categories of $60,000$ color images. We use simple data augmentation following \cite{DeeplySupervisedNet}: during training, we zero-pad $4$ pixels along each image side, and sample a $32\times 32$ region crop from the padded image or its horizontal flip; during testing, we simply use the original non-padded image. Our use of ResNet, Wide ResNet, and DenseNet for the CIFAR datasets is as follows: we use a pre-activation ResNet \cite{PreActResNet} of $68$ weight layers, whose layer specifics are given in Appendix \ref{AppendixNetArchitecture}; we use the exactly same top-performing architecture of ``WRN-28-10'' as in \cite{WideResNet}; we also use the exactly same top-performing architecture of ``DenseNet-BC'' (with the growth rate $k = 40$) as in \cite{DenseNet}. These architectures commonly aggregate features of lower layers via a top global average pooling layer, followed by a final FC layer of classification. For each network, we follow the empirical rule established in Section \ref{SecExpImageClassification} and apply CorrReg to the final FC layer. We fix $\lambda$ of CorrReg as $5\mathrm{e}^{-3}$ for all the three networks. We train ResNet and Wide ResNet for a total of $160$ epochs; learning rates are initialized as $0.1$, and decay after $80$ and $120$ epoches of training. For DenseNet, we follow \cite{DenseNet} and train for an extended duration of $300$ epochs, using mini-batches of size $64$. All other training hyperparameters are the same as described in the beginning of Section \ref{SecExp} (not necessarily the same as used in \cite{PreActResNet,WideResNet,DenseNet}). To compare with Dropout regularization, we use two schemes (denoted as Dropout-S1 and Dropout-S1S2 respectively): scheme 1 applies Dropout to (inputs of) the final FC layer of each network, which is the same as our use of CorrReg; scheme 2 follows the way in \cite{WideResNet} and \textit{additionally} applies Dropout to (inputs of) the second one of the two conv layers in each residual block of these networks. Dropping rates are tuned on the validation set with the optimal one set as $0.2$. We also try CorrReg together with the above scheme 2 (denoted as CorrReg-DropoutS2), to compare fairly with Dropout regularization. Results in Table \ref{TabExpModernNetImgClassification} show that Dropout-S1S2 improves over Dropout-S1 by providing additional regularization, especially for the CIFAR-100 dataset that contains much fewer training samples per category than CIFAR-10 does. When applying to the top FC layer alone, our proposed CorrReg consistently outperforms Dropout-S1. Moreover, the best results are obtained by CorrReg-DropoutS2 that has the combined benefit of CorrReg and Dropout regularization.

For the ImageNet dataset, we use the representative architectures of ResNet \cite{ResNet}, Wide ResNet \cite{WideResNet}, and ResNeXt \cite{ResNeXt} (more specifically, the ``ResNet-101'', and the top-performing ``WRN-50-2-bottleneck'' and ``ResNeXt-101 (64$\times$4d)'' of these architectures). For data augmentation, we adopt the same scheme as in \cite{ResNet}: during training, we randomly sample a $224\times 224$ region crop from an image or its horizontal flip; during testing, we use a single crop of size $224\times 224$. Learning rates are initialized as $0.1$, and decay by a factor of $0.1$ at $50\%$ and $75\%$ of the total $90$ training epochs, using mini-batches of size $256$.  For each network, we again apply CorrReg to the top FC layer, using a fixed $\lambda$ value of $1\mathrm{e}^{-3}$. We also apply Dropout to (inputs of) the same FC layers of these networks, where dropping rate is again set as $0.2$. Table \ref{TabExpModernNetImageNet} shows the comparative results. While Dropout regularization may not have effect on these architectures, our proposed CorrReg steadily achieves performance improvement.

\begin{table}[h]
\caption{Results of single-crop testing on the ImageNet validation set \cite{ILSVRC15} of several deep architectures with or without regularization. Results are in the format of top-1/top-5 error rates ($\%$). } \label{TabExpModernNetImageNet}
\begin{center}
\begin{tabular}{cccc}
\hline   {\scriptsize Architecture} & {\scriptsize W/O Regu.} & {\scriptsize With Dropout} & {\scriptsize With CorrReg}  \\
\hline {\scriptsize ResNet \cite{ResNet}}           & $22.39/6.25$ & $22.59/6.30$ & $\mathbf{22.14}/\mathbf{6.09}$ \\

\hline {\scriptsize Wide ResNet \cite{WideResNet}}  & $22.22/6.35$ & $22.25/6.25$ & $\mathbf{22.10}/\mathbf{6.17}$ \\

\hline {\scriptsize ResNeXt \cite{ResNeXt}}         & $20.90/5.46$ & $20.82/5.70$ & $\mathbf{20.58}/\mathbf{5.40}$ \\

\hline
\end{tabular}
\end{center}
\end{table}

\vspace{0.1cm}
\noindent\textbf{Remarks} We note that experiments in this section are not intended to compare with the best results on the benchmark image classification datasets. They are to show the efficacy of our proposed CorrReg for regularization of network training: even though input data are from the same source, intermediate features may be learned to be overfitting to view-specific patterns. CorrReg effectively regularizes network training so that the final task of classification can collaboratively benefit from all feature views.

\subsection{RGB-D Object and Scene Recognition}
\label{SecExpRGBDRecog}

In this section, we use RGB-D object dataset \cite{RGBDObjDataset} and SUN RGB-D scene dataset \cite{SUNdataset} to investigate the efficacy of CorrReg for practical problems of multi-view learning. The RGB-D object dataset contains $207,920$ RGB-D video frames of $51$ classes of $300$ object instances captured from different views, with roughly $600$ frames per object instance. We sample these videos by every $5^{th}$ frame of each video. We use the $10$ random dataset splits provided by \cite{RGBDObjDataset}, with each split containing different object instances of all the $51$ classes. For these splits, there are on average about $34,000$ images for training and $6,900$ images for testing. This dataset is intensively used for comparative studies of alternative baselines, investigation of our proposed method under different settings, and also for robustness test against contamination of input data. SUN RGB-D scene dataset is a benchmark suite for indoor scene understanding, including $10,355$ RGB-D images. For scene recognition, we follow \cite{SUNdataset} and select the $19$ most common categories, each of which has at least $80$ RGB-D images in the dataset; we then divide the data into training and test sets, giving $4,845$ images for training and $4,659$ ones for testing. For both RGB-D object and SUN RGB-D datasets, we compute surface normal (SN) \cite{upgradedHMP} from each depth image as input depth features.

\vspace{0.1cm}
\subsubsection{Comparative Studies of Alternative Baselines}
\label{SecExpRGBDRecogComp}

\begin{figure}[t]
\centering
\includegraphics[scale=0.25]{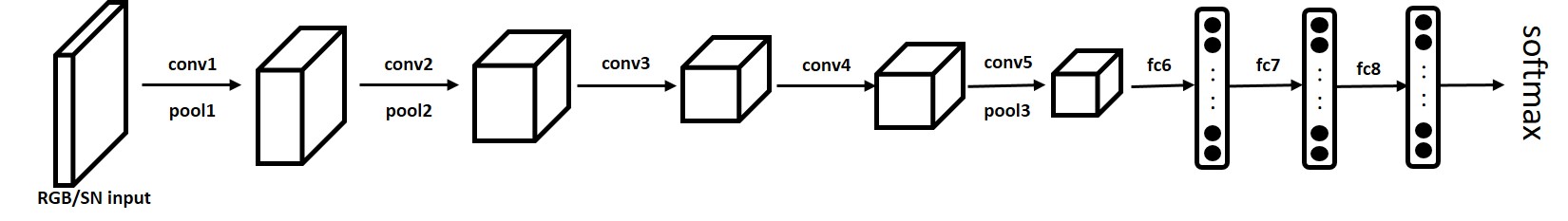} \\
\centerline{（(a)）}
\includegraphics[scale=0.25]{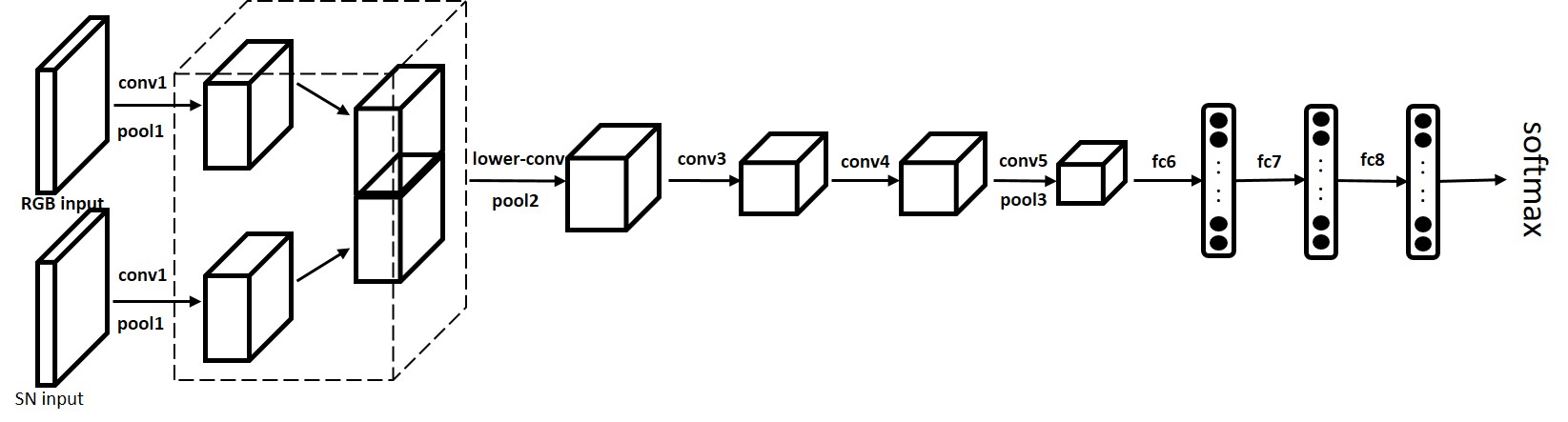} \\
\centerline{（(b)）}
\centerline{（）}
\includegraphics[scale=0.25]{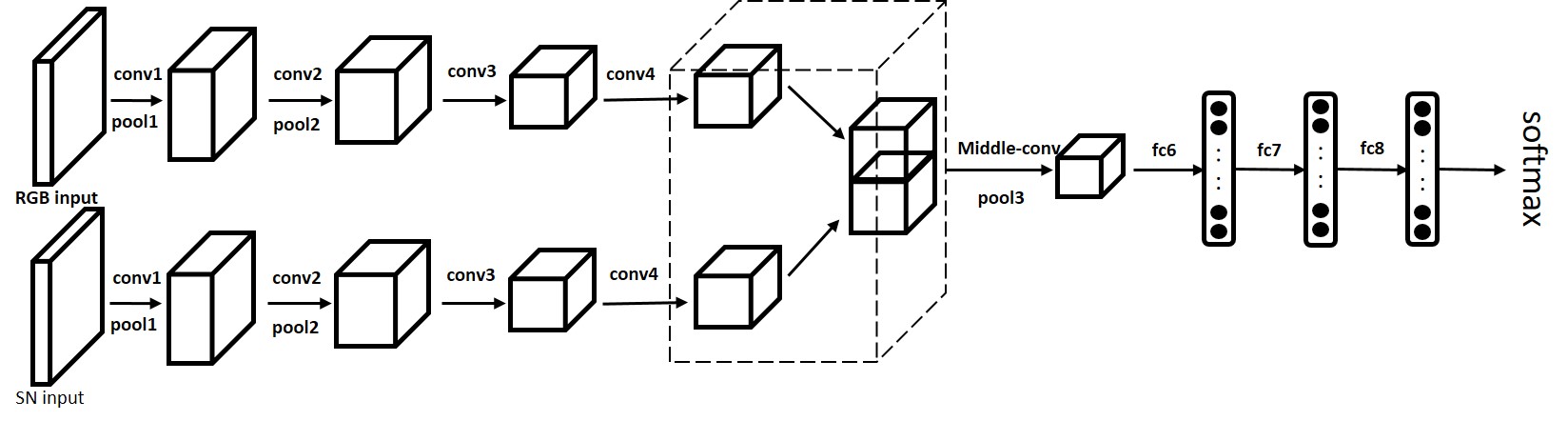} \\
\centerline{（(c)）}
\includegraphics[scale=0.25]{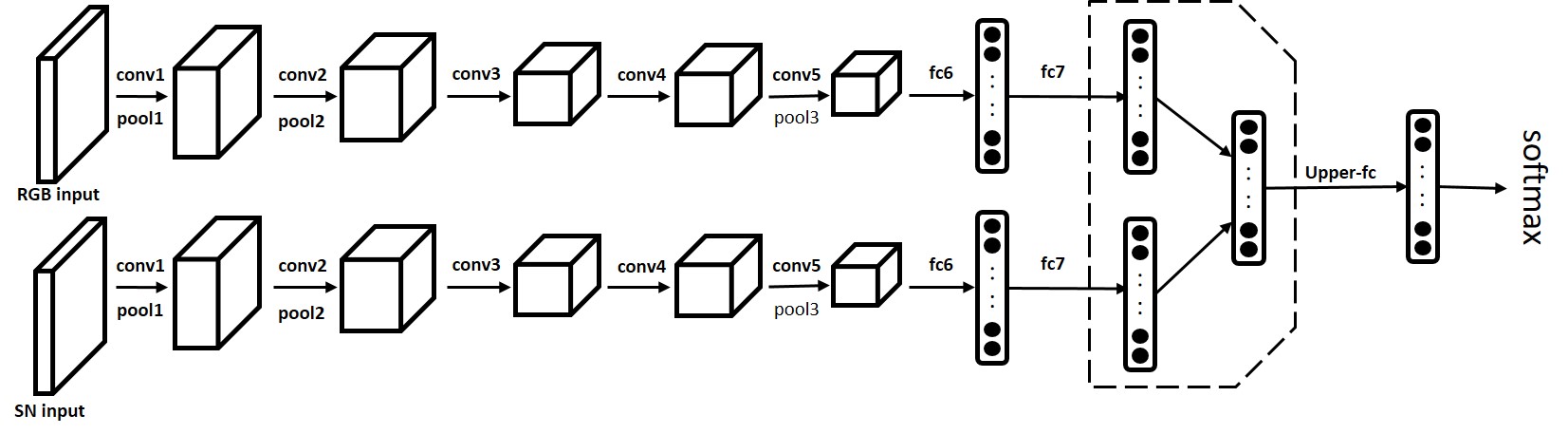} \\
\centerline{（(d)）}

\caption{Network architectures used for experiments of RGB-D object recognition. The corresponding layer parameters (numbers of feature maps, filter sizes, etc) are given in Appendix \ref{AppendixNetArchitecture}. (a) The plain (RGB or depth) ConvNets. (b)-(d) The RGB-Depth ConvNets concatenating features of the two lower streams respectively at lower, middle, or upper ``heights'' of network layers. }  \label{FigRGBDRecogAlexNet}
\end{figure}

Our comparative studies of alternative baselines for RGB-D object recognition are based on adaptations of an 8-layer ConvNet. The adaptations consist of two lower, parallel streams followed by one upper, single stream, whose model architectures are shown in Figure \ref{FigRGBDRecogAlexNet} and whose layer parameters (numbers of feature maps, filter sizes, etc) are given in Appendix \ref{AppendixNetArchitecture}. RGB and depth/SN images are respectively taken as inputs of the two lower streams, whose outputs are concatenated as inputs of the upper stream. We term such adapted networks as RGB-Depth ConvNets. In this work, we investigate the effects of concatenating features of the two lower streams at different ``heights'' (lower, middle, or upper layers) of RGB-Depth ConvNets, as illustrated in Figure \ref{FigRGBDRecogAlexNet}. Alternative to RGB-Depth ConvNets are plain ConvNets that consist of one of the two lower streams of RGB-Depth ConvNets and the upper stream, as shown in Figure \ref{FigRGBDRecogAlexNet}. Such plain ConvNets can be used for both RGB and depth images, and we term them as RGB ConvNet and Depth ConvNet respectively.

The above networks suggest several baseline methods for RGB-D object recognition. In particular, given existence of both RGB and depth images during training and test phases, one may separately train RGB ConvNet or Depth ConvNet using single-modal images, and use the trained models for single-modal inference. Alternatively, one may use the above RGB-Depth ConvNets that concatenate features of individual modalities at different heights for multi-modal training and inference. To train these baseline networks, we use common data augmentation practices on the RGB-D object dataset: we first re-scale each training image to the size of $150 \times 150$, from which or the horizontal flip of which we randomly crop a region of the size $143 \times 143$. The learning rate is initialized as $0.01$ and decays by a factor of $0.1$ when learning curves plateau. Table \ref{TabExplittlenetRGBDobjectClassification} shows that while concatenating features at lower or middle layers of RGB-Depth ConvNets is not effective, feature concatenation at an upper layer of RGB-Depth ConvNet achieves improved performance over single-modal networks.

\begin{table}[h]
\caption{Recognition accuracies($\%$) on RGB-D object dataset \cite{RGBDObjDataset} using architectures in Figure \ref{FigRGBDRecogAlexNet} and various regularization methods. Results are in the format of $\textrm{mean} \pm \textrm{standard deviation}$. } \label{TabExplittlenetRGBDobjectClassification}
\begin{center}
\begin{tabular}{ccccc}
\hline Methods                              & Accuracy         &\\
\hline

{\scriptsize RGB ConvNet}                                 & $79.83 \pm 2.06$  \\

{\scriptsize Depth Convnet}                               & $83.49 \pm 2.00$  \\

{\scriptsize RGB-Depth ConvNet}                    &                   \\
{\scriptsize (concat. at a lower height)}          & $79.48 \pm 2.64$  \\

{\scriptsize RGB-Depth ConvNet}                    &                   \\
{\scriptsize (concat. at a middle height)}         & $79.56 \pm 3.66$  \\

{\scriptsize RGB-Depth ConvNet}                     &                   \\
{\scriptsize (concat. at an upper height)}          & $87.99 \pm 1.51$  \\

\hline

{\scriptsize RGB-Depth ConvNet with Dropout}                  & $88.33 \pm 1.69$  \\

{\scriptsize RGB-Depth ConvNet with L2Regu}                  & $88.65 \pm 1.37$  \\

{\scriptsize RGB-Depth ConvNet with CorrReg}                  & $\mathbf{89.16 \pm 1.18}$  \\

{\scriptsize RGB-Depth ConvNet with L2Regu \& Dropout}       &  $88.58 \pm 1.85$  \\

{\scriptsize RGB-Depth ConvNet with CorrReg \& Dropout}     & $\mathbf{89.65 \pm 1.40}$  \\

\hline

\end{tabular}
\end{center}
\end{table}

\begin{table}[h]
\caption{Accuracies($\%$) of different $\lambda$ values when applying CorrReg to a RGB-Depth ConvNet (Figure \ref{FigRGBDRecogAlexNet}-(d)) for RGB-D object recognition \cite{RGBDObjDataset}. } \label{TableRGBDLambdaTuning}
\begin{center}
\begin{tabular}{ccccccc}
\hline $\lambda$ & $10^{-4}$ & $10^{-3}$ & $10^{-2}$ & $10^{-1}$ &$10^{0}$ & $10^{1}$  \\
\hline $\textrm{mean}$ & $88.18$    & $87.86$    & $88.08$    & $89.16$    & $89.15$    & $86.08$ \\
       $\pm \textrm{std}$ & {\scriptsize $\pm 1.22$} & {\scriptsize $\pm 1.85$} & {\scriptsize $\pm 1.26$} & {\scriptsize $\pm 1.18$} & {\scriptsize $\pm 1.93$} & {\scriptsize $\pm 1.39$} \\

\hline

\end{tabular}
\end{center}
\end{table}

\begin{table*}[ht]
\caption{Robustness test by adding random occlusion blocks of varying sizes to test RGB and depth (SN) images of the RGB-D object dataset \cite{RGBDObjDataset}. Trained networks in Section \ref{SecExpRGBDRecogComp} are used for these experiments. Results are in terms of recognition accuracy($\%$). } \vspace{-0.2cm} \label{TabExpOclussion}
\begin{center}
\begin{tabular}{|c|cc|cc|cc|cc|cc|}
\hline Occlusion size                                 && $10\times 10$ && $20\times 20$ && $30\times 30$  && $40\times 40$ && $50\times 50$  \\
\hline {\scriptsize RGB ConvNet}                                    && $79.25$       && $75.29$       && $63.52$       && $45.58$       && $28.35$ \\
\hline {\scriptsize Depth ConvNet}                                  && $83.89$       && $82.71$       && $80.07$       && $71.96$       && $54.67$ \\
\hline {\scriptsize RGB-Depth ConvNet}                              && $88.05$       && $86.81$       && $80.60$       && $66.32$       && $43.53$ \\
\hline {\scriptsize RGB-Depth ConvNet with Dropout}                 && $88.34$       && $86.98$       && $80.71$       && $66.10$       && $44.66$ \\
\hline {\scriptsize RGB-Depth ConvNet with CorrReg}                 && $89.27$       && $88.66$       && $\mathbf{85.33}$       && $\mathbf{75.73}$       && $\mathbf{56.75}$ \\
\hline {\scriptsize RGB-Depth ConvNet with CorrReg \& Dropout}      && $\mathbf{89.70}$       && $\mathbf{88.93}$       && $84.54$       && $73.28$       && $53.46$ \\

\hline
\end{tabular}
\end{center}
\end{table*}

The above baselines fuse multi-modal features via direct concatenation. Discussions in Section \ref{SecIntro} suggest that one may apply \emph{network regularization at fusion} to help collaboratively learn from multi-view features, which includes existing methods such as Dropout \cite{Dropout}, and also our proposed CorrReg that explicitly leverages multi-view learning criteria. More specifically, we apply CorrReg to the first layer of the upper stream of RGB-Depth ConvNet that concatenates multi-view features at upper ``height'', making it become a CorrReg fusion layer, or alternatively apply Dropout to inputs of the first layer of the upper stream. As noted in Section \ref{SecLiterature}, an approximate measure of correlation is used in \cite{CIM4MultiModalDL} that simply computes the (squared) Euclidean distance between features of individual views. We also consider this simple correlation measure as a baseline regularizer of (\ref{EqnReguNNObj}), and term such a method as L2Regu. We compare with L2Regu by applying it to the same first layer of the upper stream of RGB-Depth ConvNet. We set the $\lambda$ value of CorrReg as $1\mathrm{e}^{-1}$, the same penalty parameter of L2Regu as $1\mathrm{e}^{-1}$, and the dropping rate of Dropout as $0.5$, which are determined by tuning on the validation set. Results in Table \ref{TabExplittlenetRGBDobjectClassification} show that either Dropout, L2Regu, or CorrReg improves performance over that of direct concatenation, and CorrReg outperforms Dropout and L2Regu, showing the advantage of CorrReg in practical multi-view learning problems. Table \ref{TableRGBDLambdaTuning} also gives results of CorrReg when using different $\lambda$ values. To investigate whether the effect of CorrReg (or L2Regu) is complementary to that of Dropout, we also use both CorrReg (or L2Regu) and Dropout in RGB-Depth ConvNet. Using L2Regu together with Dropout may not improve over L2Regu itself; instead, using CorrReg together with Dropout further improves the performance, showing the advantage of CorrReg for complementary regularization.


\vspace{0.1cm}
\subsubsection{Robustness against Contamination of Input Data}

An important property of multi-view learning is that inference should be less influenced when input data are contaminated \cite{MVIntactSpaceLearning}. In this section, we simulate such testing scenarios by adding random occlusion blocks to input RGB and depth images. Occlusion blocks are obtained by setting pixel values of the occluded regions as $0$. We use the trained networks in Section \ref{SecExpRGBDRecogComp} for these investigations.  Table \ref{TabExpOclussion} reports comparative results under different sizes of random occlusion. Compared with RGB ConvNet and Depth ConvNet, direct feature concatenation using RGB-Depth ConvNet may not provide better robustness against contamination of input data. RGB-Depth ConvNet with our proposed CorrReg improves the robustness, and performs consistently better than plain RGB-Depth ConvNet and the one with Dropout do.

\vspace{0.1cm}
\subsubsection{Comparisons with the state of the art}
\label{SecExpRGBDRecogStateOfTheArt}

State-of-the-art results on RGB-D object recognition are obtained by using either advanced base models (e.g., ResNets \cite{ResNet}) with parameters pre-trained on ImageNet \cite{CIM4MultiModalDL}, or advanced feature encoding scheme \cite{MDSICNN}. We follow \cite{CIM4MultiModalDL} and use two ResNet-50 \cite{ResNet} (after removing its last FC layer of classifier) as the lower, parallel streams, whose $2048$-dimensional output feature vectors are concatenated as the input vector of a FC based CorrReg fusion layer, followed by the last layer of $51$-way classifier. The two lower streams respectively take RGB and SN images as inputs.  We term such a constructed network as RGB-Depth ResNet. For data augmentation, we follow \cite{WangMMSS} by first re-scaling training images to the size of $256 \times 256$, and then randomly cropping regions of the size $224 \times 224$ from them or their horizontal flips. RGB-Depth ResNet is fine-tuned with the initial learning rates of $0.0001$ for the lower streams and $0.01$ for the upper stream. We use mini-batches of size 64, and set the $\lambda$ value of CorrReg as $1\mathrm{e}^{-1}$ and the dropping rate of Dropout as $0.8$. Other training hyperparameters are the same as described in the beginning of Section \ref{SecExp}.

\begin{table}[h]
\caption{Recognition accuracies ($\%$) of different methods on the RGB-D object dataset \cite{RGBDObjDataset}. Results are in the format of $\textrm{mean} \pm \textrm{standard deviation}$. } \vspace{-0.2cm} \label{TabExpRGBDobjectClassification}
\begin{center}
\begin{tabular}{ccccc}
\hline Methods                              & Accuracy         &\\
\hline

{\scriptsize Nonlinear SVM\cite{RGBDObjDataset}}          & $83.9 \pm 3.5$    \\

{\scriptsize CKM \cite{CKM}}                              & $86.4 \pm 2.3$    \\

{\scriptsize CNN-RNN \cite{ConvRecursive3DObjRecog}}      & $86.8 \pm 3.3$    \\

{\scriptsize upgraded HMP \cite{upgradedHMP}}             & $87.5 \pm 2.9$    \\

{\scriptsize MMSS \cite{WangMMSS}}                            & $88.5 \pm 2.2$    \\

{\scriptsize Fus-CNN \cite{eitelmultimodal}}              & $91.3 \pm 1.4$    \\

{\scriptsize CIMDL-ResNet \cite{CIM4MultiModalDL}}        & $92.4 \pm 1.8$    \\

{\scriptsize MDSI-CNN \cite{MDSICNN}}                     & $92.8 \pm 1.2$    \\

\hline

{\scriptsize RGB ResNet}                                 & $90.5 \pm 1.6$  \\

{\scriptsize Depth ResNet}                               & $85.5 \pm 2.4$  \\

{\scriptsize RGB-Depth ResNet}                         & $92.5 \pm 1.2$  \\

{\scriptsize RGB-Depth ResNet with Dropout}                  & $93.1 \pm 1.4$  \\

{\scriptsize RGB-Depth ResNet with CorrReg}                  & $\mathbf{93.4 \pm 1.6}$  \\

{\scriptsize RGB-Depth ResNet with CorrReg \& Dropout}       & $\mathbf{93.6 \pm 1.6}$  \\

\hline
\end{tabular}
\end{center}
\end{table}

Note that the method \cite{CIM4MultiModalDL} uses the same ImageNet pre-trained base models (i.e., ResNet-50) as we do. It is interesting to observe in Table \ref{TabExpRGBDobjectClassification} that our result of RGB-Depth ResNet that concatenates RGB and depth features directly is better than those from most of existing methods. Regularizing RGB-Depth ResNet with either Dropout or CorrReg further improves the result, with CorrReg achieving better improvement. Using CorrReg together with Dropout achieves the new state of the art of $93.6\%$.

\vspace{0.1cm}
\subsubsection{Results on RGB-D Scene Recognition}

In this section, we report experiments of RGB-D scene recognition on the SUN RGB-D dataset \cite{SUNdataset}. We use the same network architectures and training manners as in Section \ref{SecExpRGBDRecogStateOfTheArt}, with the only difference that replaces the 51-way softmax classifiers with the 19-way ones. Results in Table \ref{TabExpSUNClassification} tell that RGB-Depth ResNet using simple feature concatenation outperforms existing methods that have complicated feature fusion schemes and/or training criteria. Regularizing RGB-Depth ResNet with CorrReg and/or Dropout further improves the result to the new state of the art.

\begin{table}[h]
\caption{Recognition accuracies ($\%$) of different methods on the SUN RGB-D scene dataset \cite{SUNdataset}. }  \label{TabExpSUNClassification}
\begin{center}
\begin{tabular}{ccccc}
\hline Methods                              & Accuracy     &\\
\hline

{\scriptsize GIST $+$ RBF Kernel SVM \cite{SUNdataset}}   & $23.0$        \\

{\scriptsize Place-CNN $+$ Linear SVM \cite{SUNdataset}}  & $37.2$        \\

{\scriptsize Place-CNN $+$ RBF Kernel SVM \cite{SUNdataset}}  & $39.0$        \\

{\scriptsize SSCNN \cite{SSCNN}}                          & $41.3$        \\

{\scriptsize DMFF \cite{DMFF}}                            & $41.5$        \\

{\scriptsize MDSI-CNN \cite{MDSICNN}}                     & $45.2$        \\

{\scriptsize FV-CNN \cite{FVCNN}}                         & $48.1$        \\

{\scriptsize RGB-D-CNN(wSVM) \cite{song2017depth}}        & $52.4$        \\

{\scriptsize BilinearCNN \cite{DBSNN}}                         & $55.5$        \\

\hline

{\scriptsize RGB ResNet}                                 & $57.8$        \\

{\scriptsize Depth ResNet}                               & $47.2$        \\

{\scriptsize RGB-Depth ResNet}                         & $59.6$        \\

{\scriptsize RGB-Depth ResNet with Dropout}                  & $60.0$        \\

{\scriptsize RGB-Depth ResNet with CorrReg}                  & $\mathbf{60.7}$        \\

{\scriptsize RGB-Depth ResNet with CorrReg \& Dropout}       & $\mathbf{61.0}$        \\

\hline
\end{tabular}
\end{center}
\end{table}

\subsection{Multi-view Recognition of 3D Object Shapes}
\label{SecExpModelNetRecog}

We conduct experiments of multi-view 3D object recognition on the ModelNet40 dataset \cite{wu20153d} to investigate the efficacy of CorrReg for practical problems with data of more than two views. The ModelNet40 dataset contains $12,311$ CAD models (meshes) of $40$ object categories, with $9,843$ models for training and $2,468$ ones for testing. To prepare images of multiple views from each object model, we follow the $1^{st}$ camera set-up in \cite{su2015multi} and assume that each model is upright oriented; $12$ virtual cameras, pointing towards the model centroid, are evenly distributed (with intervals of $30$ degrees) around a horizontal circle that is elevated $30$ degrees from the ground plane; 2D images are rendered from these $12$ camera views.

\begin{figure}[ht]
\centering
\includegraphics[scale=0.26]{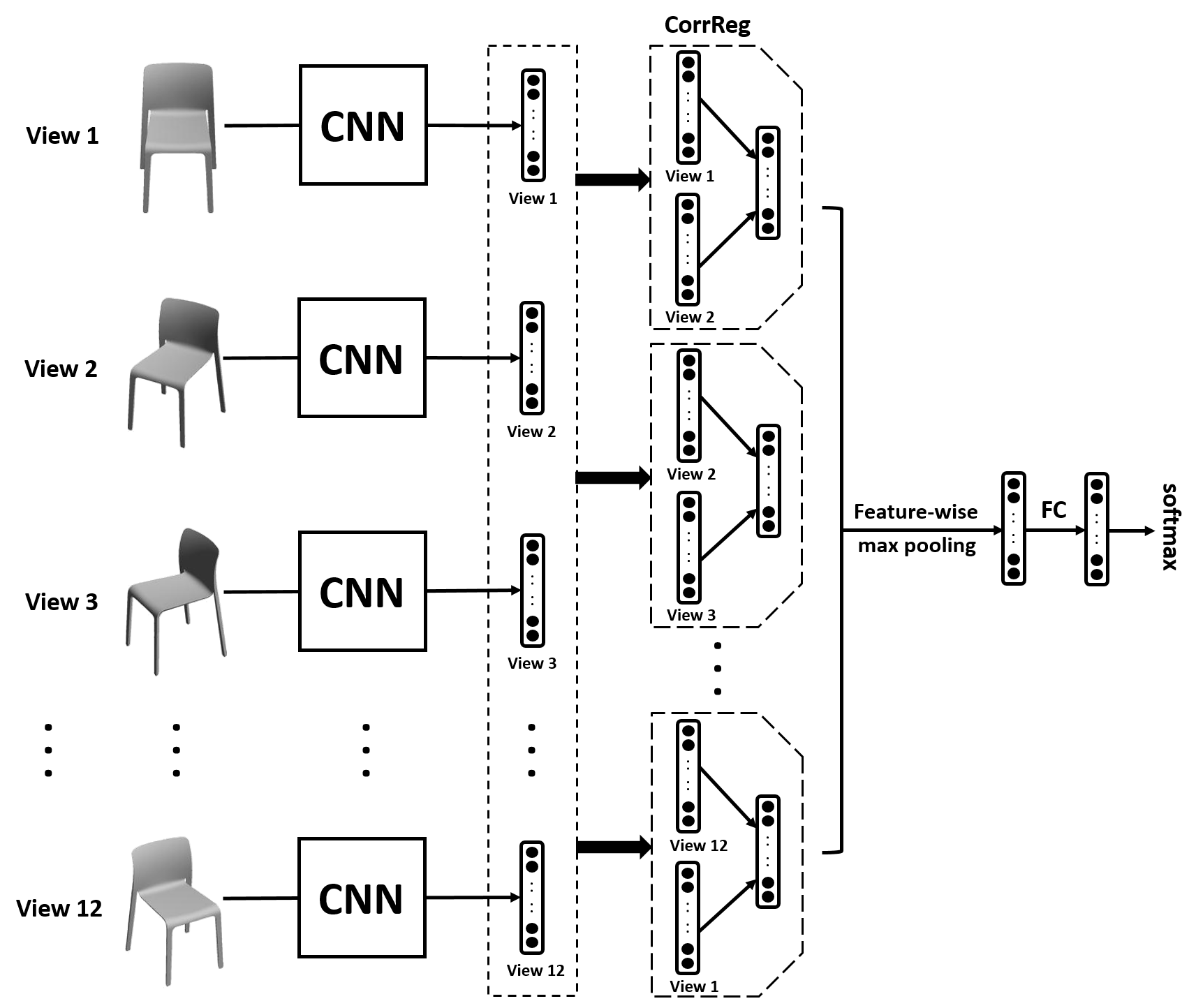} \\
\caption{Illustration of the multi-view fusion network used for experiments of multi-view 3D object recognition. Thick right arrows represent data re-organization by pairing feature vectors of individual views. }  \label{FigMV3DArch}
\end{figure}

Based on a very simple architecture of MVCNN \cite{su2015multi} for multi-view based 3D object recognition, where features of individual views extracted from lower, parallel layer streams are aggregated via feature-wise max pooling, we design a Multi-view Fusion Network (MvFusionNet) as shown in Figure \ref{FigMV3DArch}. By pairing neighboring views, MvFusionNet re-organizes feature vectors of individual views from lower streams into an equal number of pairs of feature vectors. Each of such pairs is then fed into a fusion layer where CorrReg can also be applied to form a CorrReg fusion layer. Feature-wise max pooling is subsequently applied to outputs of these fusion layers, and MvFusionNet ends with a FC layer of classifier. We investigate here whether CorrReg is helpful for feature aggregation of different views by regularizing such a constructed MvFusionNet.

Lower streams of MvFusionNet are adapted from ResNet-101 \cite{ResNet} that is pre-trained on ImageNet\cite{CIM4MultiModalDL}. To train MvFusionNet, we use a mini-batch of $16$ (i.e., $16\times 12 = 192$ images); the learning rates start at $0.001$ and decay at the rate of $0.1$ when learning curves plateau. The penalty $\lambda$ of CorrReg is set as $5\mathrm{e}^{-4}$. We report in Table \ref{TabExpMVModelNet40} results of MvFusionNet without or with CorrReg regularization, where we also compare with recent state-of-the-art results \cite{DominantSetClustering,yu2018multi,feng2018gvcnn} on ModelNet40 whose multi-view images are prepared following the same style of $1^{st}$ camera set-up in \cite{su2015multi} (i.e., $12$ camera views pointing towards upright orientation of object models). Due to varying architectural designs, network optimizers, and feature aggregation schemes, results of different methods in Table \ref{TabExpMVModelNet40} may not be directly comparable; nevertheless, it confirms the efficacy of CorrReg for better feature learning and aggregation from multiple views of 3D object shapes. We note that results of multi-view based methods on ModelNet40 depend heavily on how multi-view images are prepared by positioning virtual cameras on a sphere enclosing the object model. For example, the current best result on ModelNet40 is obtained in \cite{RotationNet} by selecting camera set-ups from a much richer set of camera positioning and viewpoints. We expect our results can also be boosted by using multi-view images rendered from these optimal camera set-ups.

\begin{table}[h]
\caption{Accuracies ($\%$) of multi-view 3D object recognition on the ModelNet40 dataset \cite{wu20153d}. All methods use the $1^{st}$ camera set-up in \cite{su2015multi} ($12$ camera views pointing towards upright oriented model). Best results of some methods and the corresponding view numbers are also quoted in parentheses. }  \label{TabExpMVModelNet40}
\begin{center}
\begin{tabular}{cl}
\hline Methods                              & Accuracy     \\
\hline

{\scriptsize MVCNN \cite{su2015multi}}   &   $89.9$ ($90.1$, 80 views)        \\

{\scriptsize Pairwise \cite{Pairwise4ModelNet}}   &   $90.7$        \\

{\scriptsize Dominant Set Clustering \cite{DominantSetClustering}}   &   $92.2$ ($93.0$, 24 views)        \\

{\scriptsize MHBN \cite{yu2018multi}}   &   $93.4$ ($94.1$, 6 views)        \\

{\scriptsize GVCNN \cite{feng2018gvcnn}}   &   $92.6$ ($93.1$, 8 views)        \\

\hline

{\scriptsize MvFusionNet}                                 & $93.9$        \\

{\scriptsize MvFusionNet with CorrReg}                    & $\mathbf{95.4}$        \\

\hline
\end{tabular}
\end{center}
\end{table}

\section{Conclusion}

We study in this paper deep multi-view learning in the context of regularized network training. We take a regularization approach via multi-view learning criteria, and propose a novel, effective, and efficient neuron-wise correlation-maximizing regularizer. We also implement such regularizers collectively as a correlation-regularized network layer (CorrReg). CorrReg can be applied to either FC or conv based fusion layers that concatenate intermediate features of individual views. Controlled experiments of benchmark image classification show that CorrReg consistently improves performance of various modern deep architectures. Applying CorrReg to multi-modal deep networks achieves the new state of the art on the benchmark RGB-D object and scene recognition datasets. In future research, we are interested in applying CorrReg to other multi-view learning problems of practical interest.

{\small
\bibliographystyle{IEEETran}
\bibliography{CorrReg}
}

\newpage

\pagestyle{empty}

\appendices

\section{Gradients of the Proposed Neuron-Wise Correlation-Maximizing Regularizer}
\label{AppendixCorrRegGradients}

We use multi-variable chain rule to derive the gradients of the neuron-wise regularizer w.r.t. the weight vectors $\frac{\partial{Corr} }{ \partial{\mathbf{w}_1} }$ and $\frac{\partial{Corr} }{ \partial{\mathbf{w}_2} }$, and also w.r.t. the input features $\frac{\partial{Corr} }{ \partial{\mathbf{x}_1^i} }$, $\frac{\partial{Corr} }{ \partial{\mathbf{x}_2^i} }$, $i = 1, \dots, m$. Their explicit forms are presented as follows (with no simplification).

\begin{equation}\label{EqnGradCorrSigma1}
\frac{\partial{Corr} }{ \partial{\sigma_1^2} } = -\frac{1}{2} (\sigma_1^2\sigma_2^2 + \epsilon)^{-\frac{3}{2}} \cdot \sigma_2^2 \cdot \sum_{i=1}^m (y_1^i - \mu_1)(y_2^i - \mu_2) \nonumber
\end{equation}

\begin{equation}\label{EqnGradCorrSigma2}
\frac{\partial{Corr} }{ \partial{\sigma_2^2} } = -\frac{1}{2} (\sigma_1^2\sigma_2^2 + \epsilon)^{-\frac{3}{2}} \cdot \sigma_1^2 \cdot \sum_{i=1}^m (y_1^i - \mu_1)(y_2^i - \mu_2) \nonumber
\end{equation}

\begin{equation}\label{EqnGradCorrMu1}
\frac{\partial{Corr} }{ \partial{\mu_1} } = \frac{ -\sum_{i=1}^m(y_2^i - \mu_2) }{ \sqrt{\sigma_1^2\sigma_2^2 + \epsilon} }  + \frac{\partial{Corr} }{ \partial{\sigma_1^2} } \cdot (-2) \sum_{i=1}^m(y_1^i - \mu_1) \nonumber
\end{equation}

\begin{equation}\label{EqnGradCorrMu2}
\frac{\partial{Corr} }{ \partial{\mu_2} } = \frac{ -\sum_{i=1}^m(y_1^i - \mu_1) }{ \sqrt{\sigma_1^2\sigma_2^2 + \epsilon} }  + \frac{\partial{Corr} }{ \partial{\sigma_2^2} } \cdot (-2) \sum_{i=1}^m(y_2^i - \mu_2) \nonumber
\end{equation}

\begin{equation}\label{EqnGradCorrY1}
\frac{\partial{Corr} }{ \partial{y_1^i} } = \frac{ (y_2^i - \mu_2) }{ \sqrt{\sigma_1^2\sigma_2^2 + \epsilon} } + \frac{\partial{Corr} }{ \partial{\mu_1} } \cdot \frac{1}{m} + \frac{\partial{Corr} }{ \partial{\sigma_1^2} } \cdot 2(y_1^i - \mu_1) \nonumber
\end{equation}

\begin{equation}\label{EqnGradCorrY2}
\frac{\partial{Corr} }{ \partial{y_2^i} } = \frac{ (y_1^i - \mu_1) }{ \sqrt{\sigma_1^2\sigma_2^2 + \epsilon} } + \frac{\partial{Corr} }{ \partial{\mu_2} } \cdot \frac{1}{m} + \frac{\partial{Corr} }{ \partial{\sigma_2^2} } \cdot 2(y_2^i - \mu_2) \nonumber
\end{equation}

\begin{equation}\label{EqnGradCorrX1X2}
\frac{\partial{Corr} }{ \partial{\mathbf{x}_1^i} } = \frac{\partial{Corr} }{ \partial{y_1^i} } \cdot \mathbf{w}_1 \ \ \ \frac{\partial{Corr} }{ \partial{\mathbf{x}_2^i} } = \frac{\partial{Corr} }{ \partial{y_2^i} } \cdot \mathbf{w}_2 \nonumber
\end{equation}

\begin{equation}\label{EqnGradCorrW1W2}
\frac{\partial{Corr} }{ \partial{\mathbf{w}_1} } = \sum_{i=1}^m \frac{\partial{Corr} }{ \partial{y_1^i} } \cdot \mathbf{x}_1^i \ \ \ \frac{\partial{Corr} }{ \partial{\mathbf{w}_2} } = \sum_{i=1}^m \frac{\partial{Corr} }{ \partial{y_2^i} } \cdot \mathbf{x}_2^i \nonumber
\end{equation}

\section{Network architectures}
\label{AppendixNetArchitecture}

Our LeNet variant in Section \ref{SecExpImageClassification} consists of $3$ conv layers (the first one is the input layer), followed by $3$ FC layers (the last one is the output layer). The first two conv layers have $32$ filters, and the third one has $64$ filters. All the three conv layers have filters of size $5\times 5$ and stride $1$. Max pooling of size $3\times 3$ is applied after the first conv layer, and average pooling of size $3\times 3$ is applied after both the second and third conv layers. The numbers of neurons for the three FC layers are respectively $256$, $64$, and $10$.

Our used ResNet in Section \ref{SecExpModernNetImgClassification} follows \cite{ResNet,PreActResNet}. In particular, we first build a ConvNet that starts with a conv layer of $16$ $3\times 3$ filters, and then sequentially stacks three types of $2X$ conv layers of $3\times 3$ filters, each of which has the feature map sizes of $32$, $16$, and $8$, and filter numbers $16$,  $32$,  and $64$, respectively. Spatial sub-sampling of feature maps is achieved by conv layers of stride $2$. The ConvNet ends with a global average pooling and FC layers, with $6X+2$ weight layers in total. Based on this ConvNet, we do (1) using an ``identify shortcut'' to connect every two conv layers of $3\times 3$ filters, and a ``projection shortcut'' when sub-sampling of feature maps is needed; (2) changing it to the pre-activation version according to \cite{PreActResNet}. We set $X = 11$ that gives $68$ weight layers.

Layer specifics of the RGB ConvNet, Depth ConvNet, and RGB-Depth ConvNets used in Section \ref{SecExpRGBDRecogComp} (Figure \ref{FigRGBDRecogAlexNet}) are presented in Table \ref{RGBDlayerparam}.

\begin{table}[]
\centering
\caption{Layer specifics of ConvNets in Figure \ref{FigRGBDRecogAlexNet}. }
\label{RGBDlayerparam}
\begin{tabular}{|c|c|}
\hline
Layer               & Filter size/Filter no./                            \\
                    &  Stride/Padding                                    \\ \hline
{\scriptsize conv1}               & {\scriptsize 7 $\times $ 7/96/2/3}                            \\ \hline
{\scriptsize conv2}               & {\scriptsize 5 $\times $ 5/96/1/2}                            \\ \hline
{\scriptsize conv3}               & {\scriptsize 3 $\times $ 3/112/1/1}                           \\ \hline
{\scriptsize conv4}               & {\scriptsize 3 $\times $ 3/128/1/1}                           \\ \hline
{\scriptsize conv5}               & {\scriptsize 3 $\times $ 3/128/1/1}                           \\ \hline
{\scriptsize fc6}                 & {\scriptsize -/1024/-/-}                                          \\ \hline
{\scriptsize fc7}                 & {\scriptsize -/512/-/-}                                           \\ \hline
{\scriptsize fc8}                 & {\scriptsize -/51/-/-}                                            \\ \hline
{\scriptsize (max) pool1, pool2, pool3} & {\scriptsize 2 $\times $ 2/-/2/1}                           \\ \hline
{\scriptsize lower-conv}          & {\scriptsize 5 $\times $ 5/192/1/2}                           \\ \hline
{\scriptsize middle-conv}         & {\scriptsize 3 $\times $ 3/256/1/1}                           \\ \hline
{\scriptsize upper-fc}            & {\scriptsize -/1024/-/-}                                          \\ \hline
\end{tabular}
\end{table}


%
%
%

\end{document}